\documentclass[a4paper]{article}
\usepackage{arxiv}
\usepackage{amssymb}
\usepackage{graphicx}
\usepackage{natbib}
\usepackage{url}
\usepackage{verbatim}

\def\vec#1{\mathchoice{\mbox{\boldmath$\displaystyle#1$}}
{\mbox{\boldmath$\textstyle#1$}}
{\mbox{\boldmath$\scriptstyle#1$}}
{\mbox{\boldmath$\scriptscriptstyle#1$}}}
\usepackage[colorinlistoftodos, textsize=tiny]{todonotes}

\usepackage[normalem]{ulem}

\usepackage{url}

\usepackage[english]{babel}
\usepackage[utf8]{inputenc}
\usepackage[T1]{fontenc}
\usepackage{amsfonts} 
\usepackage{setspace}
\usepackage{subcaption}
\usepackage{booktabs}
\usepackage{amsmath}
\usepackage{caption}
\usepackage{wrapfig}
\usepackage{array,multirow}	
\usepackage[colorlinks]{hyperref}
\usepackage[colorinlistoftodos]{todonotes}

\newcommand*\samethanks[1][\value{footnote}]{\footnotemark[#1]}

\makeatletter
\let\@fnsymbol\@arabic
\makeatother

\begin{document}
\author{Stanislaw Jastrzębski\thanks{First two authors contributed equally}\,\,\thanks{Jagiellonian University, staszek.jastrzebski@gmail.com} \thanks{MILA, Université de Montréal}\,\,, Zachary Kenton\samethanks[1]\,\,\samethanks[3]\,\,, Devansh Arpit\samethanks[3]\,\,, Nicolas Ballas\thanks{Facebook AI Research}\,\,,\\
\textbf{Asja Fischer}\thanks{Ruhr-University Bochum} \,\,, \textbf{Yoshua Bengio}\samethanks[3]\,\,\thanks{CIFAR Senior Fellow} , \textbf{Amos Storkey}\thanks{School of Informatics, University of Edinburgh.}\,\,}

\title{Three Factors Influencing Minima in SGD}
\maketitle

\begin{abstract}

We investigate the dynamical and convergent properties of stochastic gradient descent (SGD) 
applied to Deep Neural Networks (DNNs). Characterizing the relation between learning rate, batch size and the properties of the final minima, such as width or generalization, remains an open question. In order to tackle this problem we investigate the previously proposed approximation of SGD by a stochastic differential equation (SDE). {We theoretically argue that three factors - learning rate, batch size and gradient covariance - influence the minima found by SGD.
In particular we find that the ratio of learning rate to batch size is a key determinant of SGD dynamics and of the width of the final minima, and that}
 higher values of the ratio lead to wider minima and often better generalization. 
We confirm these findings experimentally. 
Further, we include experiments which show that learning rate schedules can be replaced with batch size schedules and that the ratio of learning rate to batch size is an important factor influencing the memorization process. 

\end{abstract}

\section{Introduction}
Deep neural networks (DNNs)  have demonstrated good generalization ability and achieved state-of-the-art performance in many application domains. This is despite being massively over-parameterized, and despite the fact that modern neural networks are capable of getting near zero error on the training dataset \cite{zhang2016understanding}. The reason for their success at generalization remains an open question. 

The standard way of training DNNs involves minimizing a loss function using stochastic gradient descent (SGD) or one of its variants~\cite{bottou1998online}. 
Since the loss functions of DNNs are typically non-convex functions of the parameters, with complex structure and potentially multiple minima and saddle points, SGD generally converges to different regions of parameter space, with different geometries and generalization properties, depending on optimization hyper-parameters and initialization.

Recently, several works~\cite{pmlr-v70-arpit17a,advani2017high,2016arXiv160904836S} have investigated how SGD impacts on generalization in DNNs. It has been argued that wide minima tend to generalize better than sharp ones~\cite{hochreiter1997flat,2016arXiv160904836S}.
One paper \cite{2016arXiv160904836S} empirically showed that a larger batch size correlates with sharper minima and worse generalization performance.  On the other hand, \cite{2017arXiv170304933D} discuss the existence of sharp minima which behave similarly in terms of predictions compared with wide minima. 
We argue that, even though sharp minima that have similar performance exist,
SGD does not target them. Instead it tends to find wider minima at higher noise levels in gradients and it seems to be that such wide minima found by SGD correlate with better generalization.

In this paper we find that the critical control parameter for SGD is not the batch size alone, but the ratio of the learning rate (LR) to batch size (BS), i.e. LR/BS. SGD performs similarly for different batch sizes but a constant LR/BS. On the other hand higher values for LR/BS result in convergence to wider minima, which indeed seem to result in better generalization.

Our main contributions are as follows:
\begin{itemize}
\item We note that any SGD processes with the same LR/BS are discretizations of the same Stochastic Differential Equation. 
\item We derive a relation between LR/BS and the width of the minimum found by SGD.
\item We verify experimentally that the dynamics are similar under rescaling of the LR and BS by the same amount. In particular, we investigate changing batch size, instead of learning rate, during training.
\item We verify experimentally that a larger LR/BS correlates with a wider endpoint of SGD and better generalization.
\end{itemize}

\section{Theory}
\label{sec:theory}

Let us consider a model parameterized by $\vec \theta$ where the components are $\theta_i$ for $i \in \{1, \dots, q\}$, and $q$ denotes the number of parameters. For $N$ training examples $\vec x_n, n \in \{1, ..., N\}$, we define the loss function, $L(\vec{\theta})=\frac{1}{N} \sum_{n=1}^N l(\vec{\theta}, \vec{x}_n)$, and the corresponding gradient $\mathbf{g}(\vec{\theta}) = \frac{\partial L}{\partial \vec{\theta}}$, based on the sum over the loss values for {\it all} training examples.

Stochastic gradients $\mathbf{g}^{(S)}(\vec{\theta})$ arise when we consider a minibatch $\mathcal B$ of size $S<N$ of random indices drawn uniformly from $\{1,...,N\}$
and form an (unbiased) estimate of
the gradient based on the corresponding subset of training examples
$ \mathbf{g}^{(S)}(\vec{\theta}) = 
\frac{1}{S} \sum_{n \in \mathcal B} \frac{\partial}{\partial \vec{\theta}} l(\vec{\theta}, \vec{x}_n)$.

We consider stochastic gradient descent 
 with learning rate $\eta$, as defined by the update rule 
\begin{align}
\vec{\theta}_{k+1} =\vec{\theta}_{k} - \eta\vec{g}^{(S)}(\vec{\theta}_k) \enspace, \label{eqn:sgddefn}
\end{align}
where $k$ indexes the discrete update steps.

\subsection{SGD dynamics are determined by learning rate to batch size ratio}
In this section we consider SGD as a discretization of a stochastic differential equation (SDE); in this underlying SDE, the learning rate and batch size only appear as the ratio LR/BS. 
In contrast to previous work (see related work, in Section~\ref{sec:related}, e.g. \cite{Mandt2017StochasticGD,pmlr-v70-li17f}), {we draw attention to the fact} that  SGD processes with different learning rates and batch sizes but the same ratio of learning rate to batch size are different discretizations of the same underlying SGD, and hence their dynamics are the same, as long as the discretization approximation is justified.

\textbf{Stochastic Gradient Descent:\ } We focus on SGD in the context of large datasets. 
Consider the loss gradient at a randomly chosen data point, 
\begin{equation}
\vec{g}_n(\vec{\theta}) = \frac{\partial}{\partial \vec{\theta}} l(\vec{\theta}, \vec{x}_n). \label{eqn:gradelement}
\end{equation}
Viewed as a random variable induced by the random sampling of the data items, $\vec{g}_n(\vec{\theta})$
is an unbiased estimator of the gradient $\vec{g}(\vec{\theta})$. For typical loss functions this estimator has finite covariance which we denote by $\mathbf{C}(\vec{\theta})$. In the limit of a sufficiently large dataset, each item in a batch is an independent and identically distributed (IID) sample of this estimator.

For a sufficiently large batch size $\vec{g}^{(S)}(\vec{\theta})$ is a mean of components of the form, $\vec{g}_n(\vec{\theta})$, each IID. Hence, under the central limit theorem, $\vec{g}^{(S)}(\vec{\theta})$ is approximately Gaussian with mean $g(\vec{\theta})$ and variance 
$\mathbf{\Sigma}(\vec{\theta})=(1/S) \mathbf{C}(\vec{\theta})$.

Stochastic gradient descent (\ref{eqn:sgddefn}) can be written as
\begin{align}
\vec{\theta}_{k+1} =\vec{\theta}_{k} - \eta\vec{g}(\vec{\theta}_k) + \eta (\vec{g^{(S)}}(\vec{\theta}_k) - \vec{g}(\vec{\theta}_k)) 
\enspace, \label{eqn:sgdasstochastic}
\end{align}
where we have established that $(\vec{g^{(S)}}(\vec{\theta}_k) - \vec{g}(\vec{\theta}_k))$ is an additive zero mean Gaussian random noise with variance $\mathbf{\Sigma}(\vec{\theta})=(1/S)\mathbf{C}(\vec{\theta})$. Hence we can rewrite
(\ref{eqn:sgdasstochastic}) as
\begin{align}
\vec{\theta}_{k+1} =\vec{\theta}_{k} - \eta\vec{g}(\vec{\theta}_k) + \frac{\eta}{\sqrt{S}} \vec{\epsilon}
\enspace, \label{eqn:sgdasdiscretenoisy}
\end{align}
where $\vec{\epsilon}$ is a zero mean Gaussian random variable with covariance $\mathbf{C}(\vec{\theta})$.

\textbf{Stochastic Differential Equation:\ } Consider now a stochastic differential equation (SDE)
\footnote{See \cite{Mandt2017StochasticGD} for a different SDE but which also has a discretization equivalent to SGD.}
of the form
\begin{align}
d \vec{\theta} = - \mathbf{g}( \vec{\theta}) dt + \sqrt{\frac{\eta}{S}}\mathbf{R}(\vec{\theta})d \vec{W}(t) \enspace, \label{eqn:sdeform}
\end{align}
where $\mathbf{R}(\vec{\theta})\mathbf{R}(\vec{\theta})^T=\mathbf{C}(\vec{\theta})$. In particular we use $\mathbf{R}(\mathbf{\theta}) = \mathbf{U}(\mathbf{\theta})\mathbf{D(\mathbf{\theta})}^{\frac{1}{2}}$, and the eigendecomposition of  $\mathbf{C(\mathbf{\theta})}$ is given by $\mathbf{C(\mathbf{\theta})} = \mathbf{U(\mathbf{\theta})}\mathbf{\Lambda(\mathbf{\theta})}\mathbf{U(\mathbf{\theta})}^T$, for $\mathbf{\Lambda(\mathbf{\theta})}$ the diagonal matrix of eigenvalues and 
$\mathbf{U(\mathbf{\theta})}$ the orthonormal matrix of eigenvectors of 
$\mathbf{C(\mathbf{\theta})}$.
This SDE can be discretized using the Euler-Maruyama (EuM) method\footnote{See e.g. \cite{kloeden}.} with stepsize $\eta$ to obtain precisely the same equation as \eqref{eqn:sgdasdiscretenoisy}.

Hence we can say that SGD implements an EuM approximation\footnote{For a more formal analysis, not requiring central limit arguments, see an alternative approach \cite{pmlr-v70-li17f} which also considers SGD as a discretization of an SDE. Note that the {learning rate to batch size ratio} is not present there.}
to the SDE (\ref{eqn:sdeform}). Specifically we note that in the underlying SDE the learning rate and batch size only appear in the ratio $(\eta/S)$, which we also refer to as the stochastic noise. This implies that these are not independent variables in SGD. Rather it is only their ratio that affects the path properties of the optimization process. The only independent effect of the learning rate $\eta$ is to control the stepsize of the EuM method approximation, affecting only the per batch speed at which the discrete process follows the dynamics of the SDE. There are, however, more batches in an epoch for smaller batch sizes, so the per data-point speed is the same. 

Further, when plotted on an epoch time axis, the dynamics will approximately match if the learning rate to batch size ratio is the same. This can be seen as follows: rescale both learning rate and batch size by the same amount, $\eta' = a\eta$ and $S' = a S$, for some $a>0$. Note that the different discretization stepsizes give a relation between iteration numbers, $k' = k/a$. Note also that the epoch number $e$, is related to the iteration number through $e = kS/N$. This gives $e' = k'S'/N = (k/a) (Sa)/N = kS/N = e$, i.e. the dynamics match on an epoch time axis.

\textbf{Relaxing the Central Limit Theorem Assumption:\ }
We note here as an aside that above we provided an intuitive analysis in terms of a central limit theorem argument but this can be relaxed by taking a more formal analysis through computing 
\begin{align}
\Sigma(\vec{\theta}) = \left(\frac{1}{S} - \frac{1}{N}\right)\mathbf{K}(\vec{\theta})
\end{align}
where $\mathbf{K}(\vec{\theta})$ is the sample covariance matrix
\begin{align}
\mathbf{K}(\vec{\theta}) \equiv \frac{1}{N-1}\sum_{n=1}^N \left(\vec{g}_n(\vec{\theta}) - \vec{g}(\vec{\theta})\right)
\left(\vec{g}_n(\vec{\theta}) - \vec{g}(\vec{\theta})\right)^T. \label{KapproxC}
\end{align}
This was shown in e.g. \citep{2017arXiv170507562J}, but a similar result was also found earlier in \citep{2017arXiv170508741H}. By taking the limit of a small batch size compared to training set size, $S \ll N$, one achieves $\Sigma(\vec{\theta}) \approx \mathbf{K}(\vec{\theta})/S$, which is the same as the central limit theorem assumption, with the sample covariance matrix $\mathbf{K}(\vec{\theta})$ approximating $\mathbf{C}(\vec{\theta})$, the covariance of $\vec{g}_n(\vec{\theta})$.

\subsection{Learning rate to batch size ratio and the trace of the Hessian}
\label{sec:width}

{We argue in this paper that there is a theoretical relationship between the expected loss value, the level of stochastic noise in SGD ($\eta/S$) and {the width of the minima explored at this final stage of training.}  We derive that relationship in this section. }

In talking about the width of a minima, we will define it in terms of  the trace of the Hessian at the minima, $Tr(\mathbf{H})$, with a lower value of $Tr(\mathbf{H})$, the wider the minima. In order to derive the required relationship, we will make the following assumptions in the final phase of training:

\begin{description}
\item[Assumption 1] 
As we expect the training to have arrived in a local minima, the loss surface can be approximated by a quadratic bowl, with minimum at zero loss (reflecting the ability of networks to fully fit the training data). Given this
the training can be approximated by an Ornstein-Unhlenbeck process. This is a similar assumption to previous papers \cite{Mandt2017StochasticGD,2018arXiv180100173P}. 

	\item[Assumption 2] The covariance of the gradients and the Hessian of the loss approximation are approximately equal, i.e. we can sufficiently assume~$ \mathbf{C} = \mathbf{H}$. A closeness of the Hessian and the covariance of the gradients in practical training of DNNs has been argued before \cite{2017arXiv170604454S,2018arXiv180300195Z}. In Appendix~\ref{app:CasH} we discuss conditions under which $\mathbf{C} = \mathbf{H}$.
\end{description}

{The second assumption is inspired by the recently proposed explanation by \cite{2018arXiv180300195Z,DBLP:journals/corr/abs-1803-01927} for mechanism behind escaping the sharp minima, where it is proposed that SGD escapes sharp minima due to the covariance of the gradients $\mathbf{C}$ being anisotropic and aligned with the structure of the Hessian.} 

Based on Assumptions 1 and 2, the Hessian 
is positive definite, and matches the covariance $\mathbf{C}$. Hence its eigendecomposition is $\mathbf{H} = \mathbf{C} = \mathbf{V}\mathbf{\Lambda}\mathbf{V}^T$, with $\mathbf{\Lambda}$ being the diagonal matrix of positive eigenvalues, and $\mathbf{V}$ an orthonormal matrix. We can reparameterize {the model} in terms of a new variable $\vec{z}$ defined by
$\vec{z} \equiv \mathbf{V}^T(\vec{\theta}- \vec{\theta}_*)$
where $\vec{\theta}_*$ are the parameters at the minimum.
 
Starting from the SDE \eqref{eqn:sdeform}, and making the quadratic approximation of the loss  {$L(\mathbf{\theta}) \approx (\vec{\theta}- \vec{\theta}_*)^T \mathbf{H} (\vec{\theta}- \vec{\theta}_*)$} and the change of variables, results in an Ornstein-Unhlenbeck (OU) process for $\vec{z}$
\begin{align}
d \vec{z} = - \mathbf{\Lambda} \vec{z} dt + \sqrt[]{\frac{\eta }{S}} \mathbf{\Lambda}^{1/2} d\mathbf{W}(t)\enspace. \label{ouzeromean}
\end{align}

It is a standard result that the stationary distribution of an OU process of the form \eqref{ouzeromean} is Gaussian with zero mean and covariance 
$
\mathrm{cov}(\vec{z}) = \mathbb{E}(\vec{z}\vec{z}^T) = \frac{\eta}{2S}\mathbf{I} 
$.

Moreover, in terms of the new parameters {$\vec{z}$}, the expected loss can be 
written as
\begin{align}
\mathbb{E}(L) = \frac{1}{2}\sum_{i=1}^q \lambda_i \mathbb{E}(z_i^2)=\frac{\eta}{4S}\mathrm{Tr}(\mathbf{\Lambda})=\frac{\eta}{4S}\mathrm{Tr}(\mathbf{H})\ 
\enspace
\label{loss}
\end{align}
where {the expectation is over the stationary distribution of the OU process}, and the second equality follows from the expression for the OU covariance.

We see from 
{Eq. \eqref{loss}}
that the learning rate to batch size ratio {determines} the trade-off between width and expected loss associated with SGD dynamics within a minimum centred at a point of zero loss, with $\frac{\mathbb{E}(L)}{Tr(\mathbf{H} )}\propto \frac{\eta }{S}$. In the experiments which follow, we {compare} geometrical properties of minima {with}  {similar} loss values ({but different generalization properties}) {to empirically analyze this} relationship between $Tr(\mathbf{H} )$ and $\frac{\eta }{S}$.

\subsection{Special Case of Isotropic Covariance}\label{sec:equilibrium}

In this section, we look at a case in which the assumptions are different to the ones in the previous section. We will not make assumptions 1 and 2. We will instead take the limit of vanishingly small learning rate to batch size ratio, {assume} an isotropic covariance matrix, and allow the process to continue for exponentially long times to reach equilibrium.

While these assumptions are too restrictive for the analysis to be directly transferable to the practical training of DNNs, the {resulting investigations} are mathematically interesting and provide further evidence that the learning rate to batch size ratio is theoretically important in SGD.

{Let us now assume the individual gradient covariance to be isotropic, that is} $\mathbf{C}(\vec{\theta}) = \sigma^2 \mathbf{I}$, for  constant $\sigma$. 
In this special case the SDE is well known to have an analytic equilibrium distribution\footnote{The equilibrium solution is the very late time stationary (time-independent) solution with detailed balance, which means that in
the stationary solution, each individual transition balances precisely with its time reverse, resulting
in zero probability currents, see \S5.3.5 of \cite{gardiner2010stochastic}.},
given by the Gibbs-Boltzmann distribution\footnote{The Boltzmann equilibrium distribution of an SDE related to SGD was also investigated by \cite{HESKES1993199} but only for the online setting (i.e.~a batch size of one).} (see for example Section 11.4 of \cite{van1992stochastic}) 
\begin{align}
P(\vec{\theta}) & = {P_0}\exp\left({-\frac{L(\vec{\theta})}{2T}} \right) \enspace, \label{boltzmann}
\\
\text{with}\;\;\; T &\equiv \frac{\eta \sigma^2}{S}  \enspace,
\end{align}
where we have used the symbol $T$ in analogy with the temperature in a thermodynamical setting, and $P_0$ is the normalization constant\footnote{Here we assume a weak regularity condition that either the weights belong to a compact space or that the loss grows unbounded as the weights tend to infinity, e.g. by including an L2 regularization in the loss.}. If we run SGD for long enough then it will begin to sample from this equilibrium distribution. 
{By inspection of equation} \eqref{boltzmann}, we note that at higher values of 
$\eta/S$
the distribution becomes more spread out, increasing the covariance of $\vec{\theta}$, in line with our general findings of Section~\ref{sec:width}. 

We consider a setup with two minima, $A$ and $B$, and ask
which region SGD will most likely end up in.
{Starting} from the equilibrium distribution in equation \eqref{boltzmann}, we now consider the ratio of the probabilities $p_A$, $p_B$ of ending up in minima $A, B$, respectively. We characterize {the two minimia} regions by the loss values, $L_A$ and $L_B$, and the Hessians $\mathbf{H}_A$ {and $\mathbf{H}_B$, respectively}. We take a Laplace approximation\footnote{
The Laplace approximation is a common approach used to approximate integrals, used for example by \cite{6796869} and \cite{doi:10.1080/01621459.1995.10476572}. 
For a minimum  $x_0\in \Omega$ of $f$ with Hessian $H(x_0)$, the Laplace approximation is $\int_\Omega e^{Mf(x)} dx = {\left(\frac{2\pi}{M\det H(x_0)}\right)^{d/2}}e^{Mf(x_0)}$ as $M \to \infty$. }  around each minimum to evaluate the integral of the Gibbs-Boltzmann distribution around each minimum, giving the ratio of probabilities, in the limit of small temperature,
\begin{align}
\frac{{p}_A}{{p}_B} &=
\sqrt{\frac{\det{\mathbf{H}_B}}{\det{\mathbf{H}_A}}}
\exp\left( \frac{1}{2T} \left(L_B - L_A \right)\right).\label{paoverpb}
\end{align}
The first term in \eqref{paoverpb} is the ratio of the Hessian determinants, and is set by the width (volume) of the two minima. The second term is an exponent of the difference in the loss values divided by the temperature.
We see that a higher temperature decreases the influence of the exponent, so that the width of a minimum becomes more important relative to its depth, with wider minima favoured by a higher temperature. For a given $\sigma$, the temperature is controlled by the ratio of learning rate to batch size. 

\section{Experiments}\label{sec:exper}
We now present 
{an empirical analysis}
motivated by the theory discussed in the previous section.

\subsection{Learning dynamics of SGD depend on LR/BS}
In this section we look experimentally at the approximation of SGD as an SDE given in Eq.~\eqref{eqn:sdeform}, investigating how the dynamics are affected by the learning rate to batch size ratio.

\begin{figure}

\centering
 \includegraphics[width=0.95\columnwidth,trim=0.in 0.in 0.in 0.in,clip]{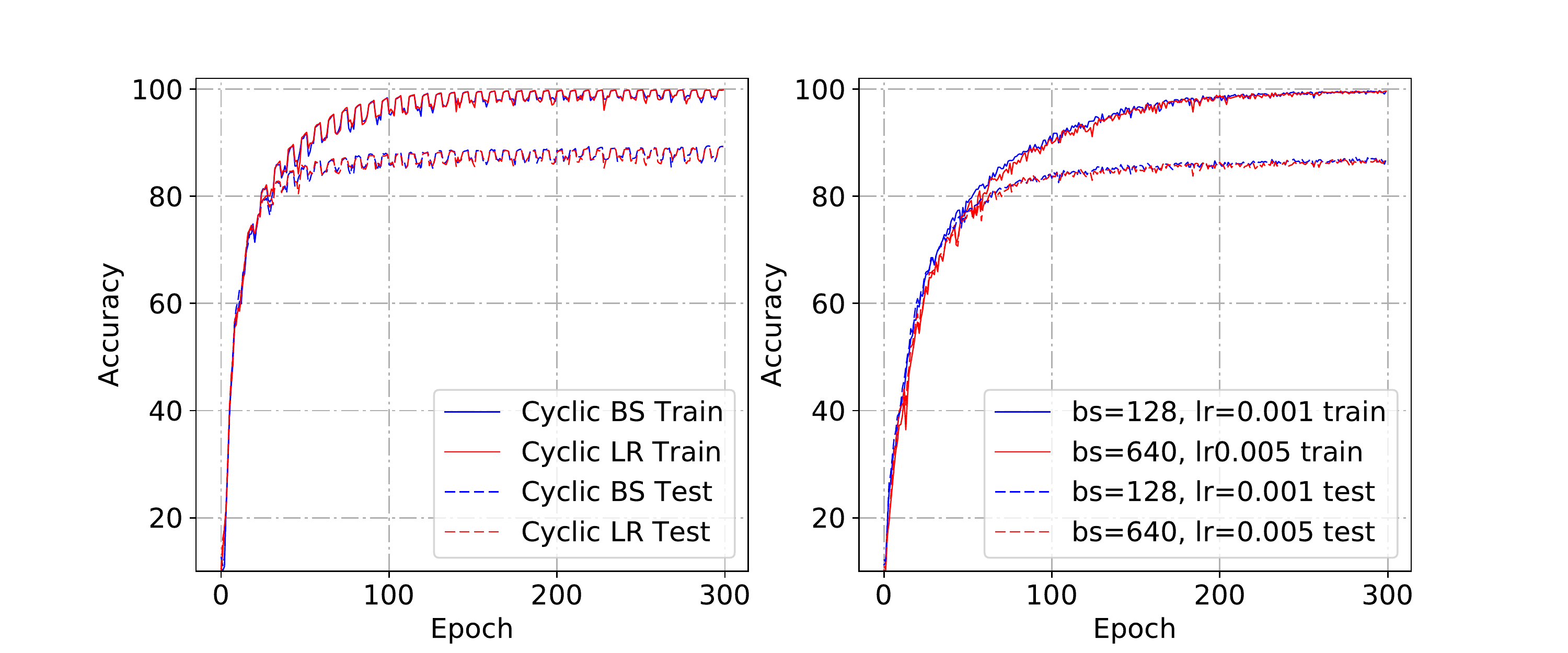}
 \caption{VGG11 on CIFAR10. Left: cyclic schedules. Right: constant $\eta$, $S$. Red and blue curves match implies dynamics set by ratio of learning rate to batch size.}
\label{fig:lr_bs_exchang}

\end{figure}

\begin{figure}
 \centering
\begin{subfigure}[t]{0.45\columnwidth}
  \centering
  \includegraphics[width=\columnwidth ,trim=0.in 0.in 0.in 0.in,clip]{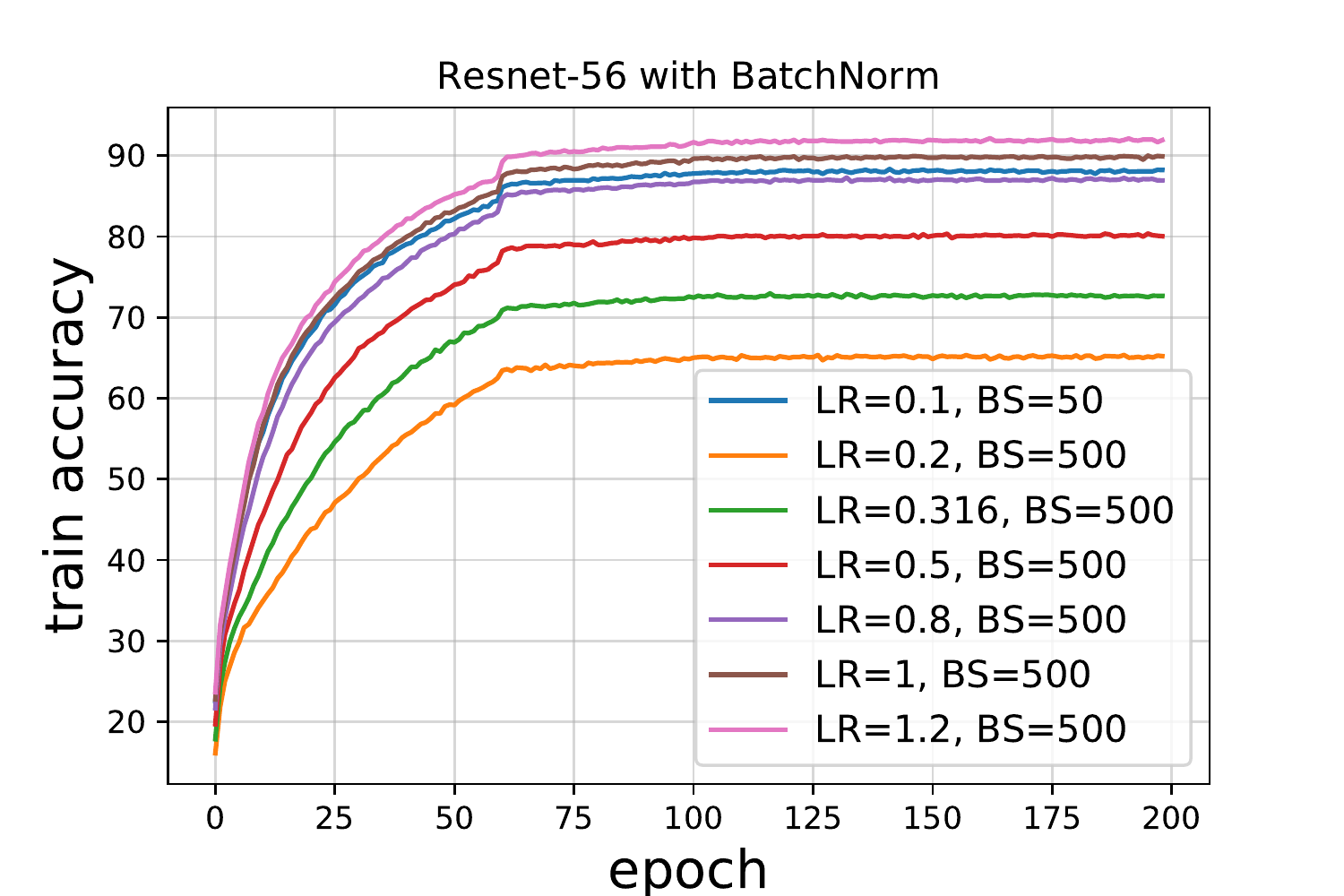}
\end{subfigure}
~
\begin{subfigure}[t]{0.45\columnwidth}
  \centering
  \includegraphics[width=\columnwidth ,trim=0.in 0.in 0.in 0.in,clip]{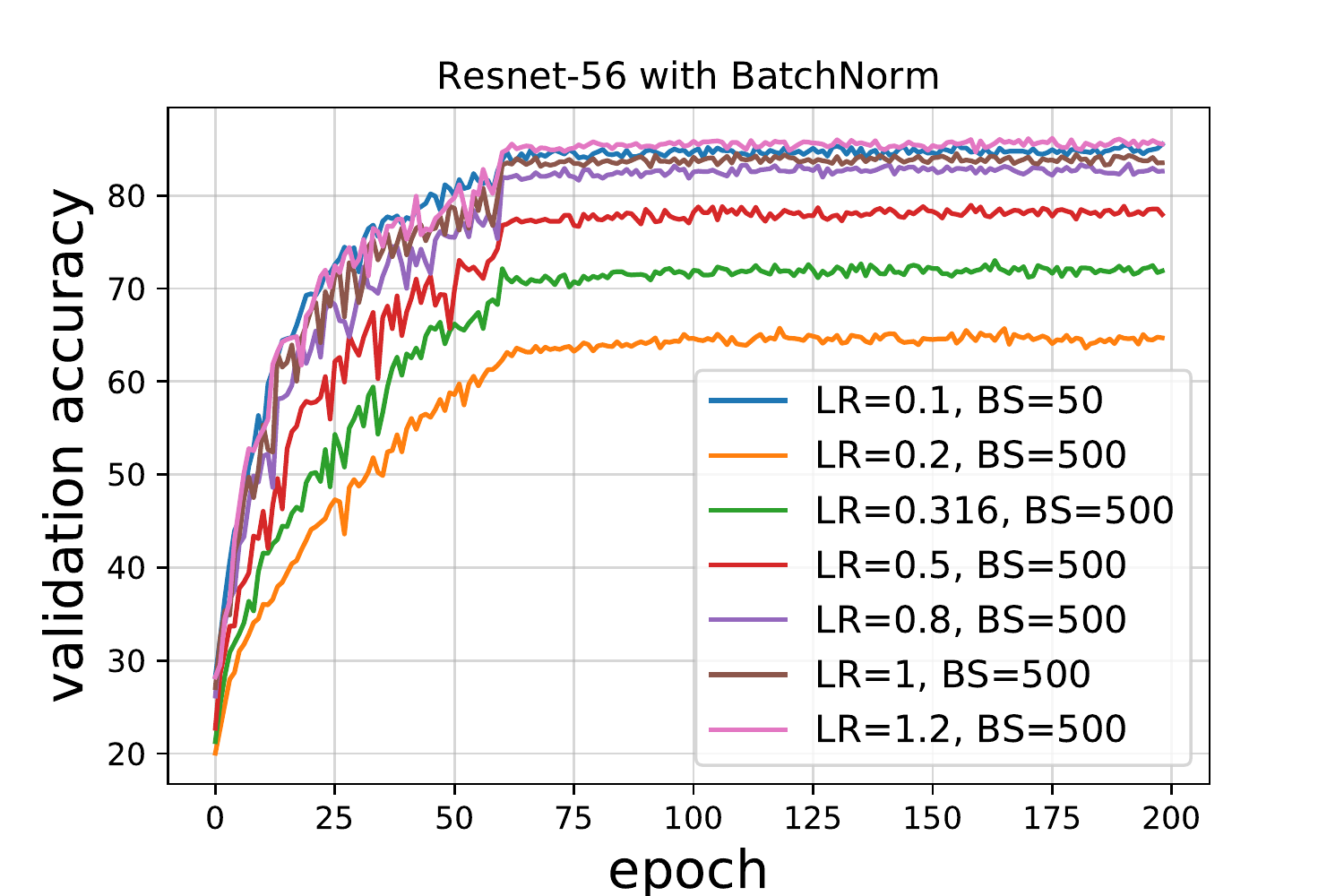}
\end{subfigure}
\vfill
\begin{subfigure}[t]{0.45\columnwidth}
  \centering
  \includegraphics[width=\columnwidth ,trim=0.in 0.in 0.in 0.in,clip]{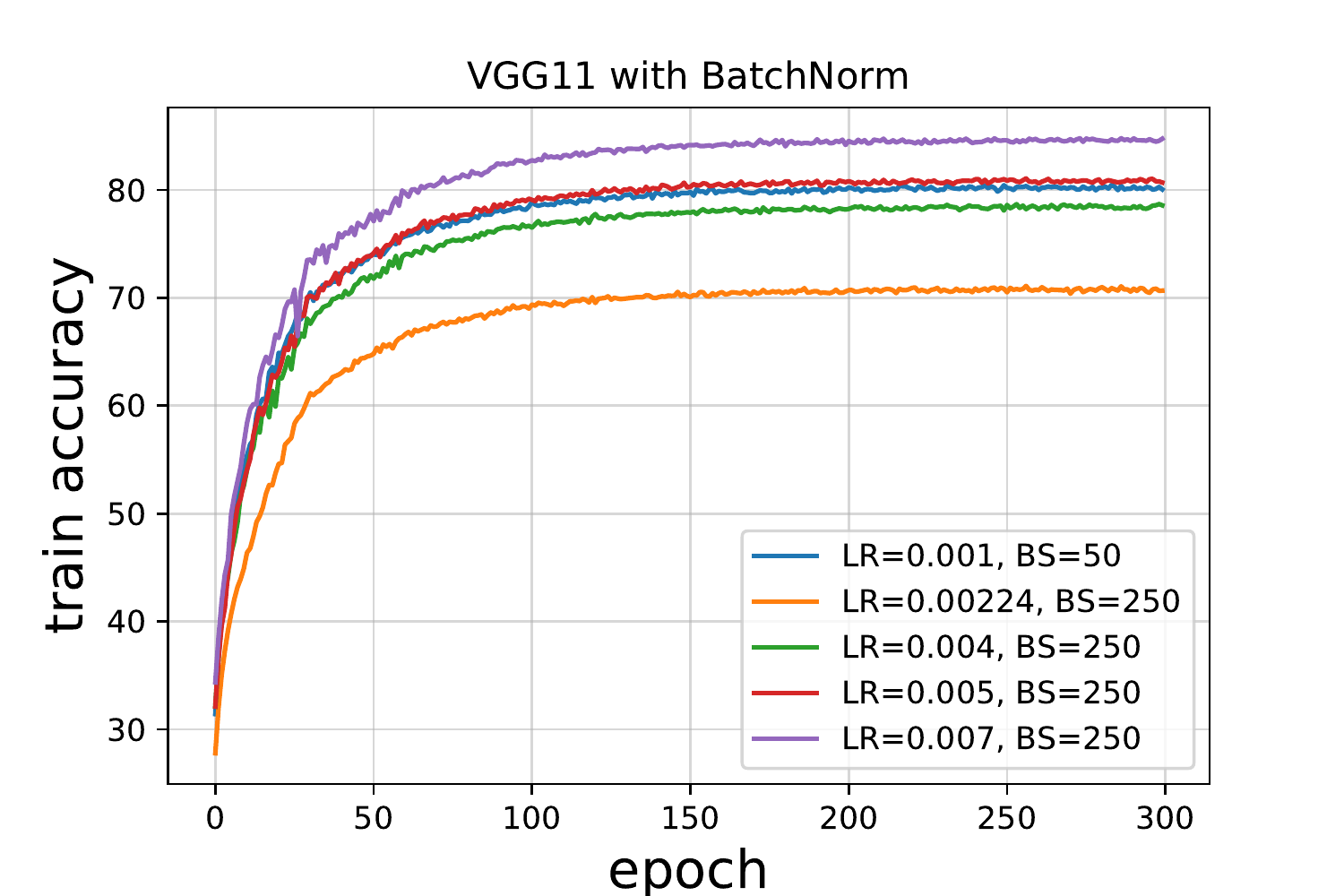}
\end{subfigure}
~
\begin{subfigure}[t]{0.45\columnwidth}
  \centering
  \includegraphics[width=\columnwidth ,trim=0.in 0.in 0.in 0.in,clip]{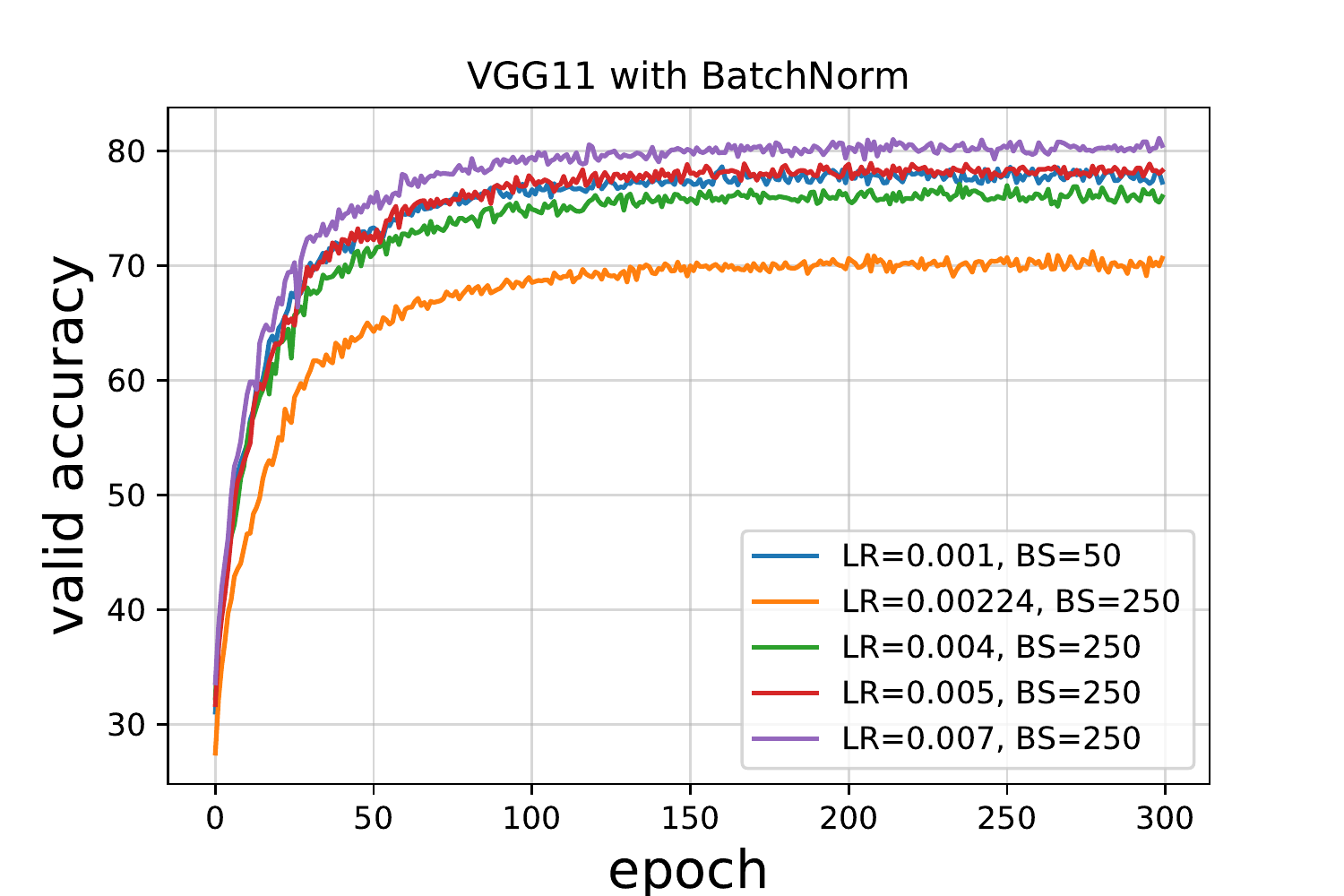}
\end{subfigure}
\caption{{ResNet (top) and VGG11 (bottom) on CIFAR10. Rescaling the learning rate to reproduce a similar learning curve when going from a small batch size (blue) to a large one. In both experiments rescaling learning rate by same amount as batch size gives a closer match than rescaling by the square root of batch size.}}
\label{fig:lr_strategies}
\end{figure}

We first look at the results of four experiments involving the VGG11 architecture\footnote{{we have adapted the final fully-connected layers of the VGG11 to be FC-512, FC-512, FC-10 so that it is compatible with the CIFAR10 dataset.} } \cite{simonyan2014very} on the CIFAR10 dataset, shown in Fig~\ref{fig:lr_bs_exchang}\footnote{
{Each experiment was repeated for 5 different random initializations.}
}.
The left plot compares two experimental settings: a cyclic batch size (CBS) schedule (blue) oscillating between 128 and 640 at fixed learning rate $\eta=0.005$, compared to {a} cyclic learning rate (CLR) schedule (red) oscillating between 0.001 to 0.005 with {a} fixed batch size {of} $S=128$. The right plot compares the results for two other experimental settings: {a} constant learning rate to batch size ratio of {$\frac{\eta}{S}=\frac{0.001}{128}$ (blue) {versus} 
$\frac{\eta}{S}=\frac{0.005}{640}$} (red).
We emphasize the similarity of the curves for each pair of experiments,
{demonstrating that} the learning dynamics are approximately invariant under changes in learning rate or batch size that keep the ratio $\eta/S$ constant.

We next ran experiments with other rescalings of the learning rate when going from a small batch size to a large one, to compare them against rescaling the learning rate exactly with the batch size.
In Fig.~\ref{fig:lr_strategies} we show the results from two experiments on
ResNet56 
and VGG11,
both trained with SGD and batch normalization on CIFAR10. In {both settings} the blue line {corresponds to training with} a small batch size of 50 and a small starting learning rate\footnote{{We used a adaptive learning rate schedule} with $\eta$ dropping by a factor of 10 on epochs 60, 100, 140, 180 for ResNet56 and by a factor of 2 every 25 epochs for VGG11.}. 
{The other lines correspond to models trained with different learning rates and a larger batch size. It becomes visible that when rescaling $\eta$ by the same amount as $S$ (brown curve for ResNet, red for VGG11) the learning curve matches fairly closely the blue curve. Other rescaling strategies such as keeping the ratio $\eta/\sqrt{S}$ constant, as suggested by \cite{2017arXiv170508741H}, (green curve for ResNet, orange for VGG) lead to larger differences in the learning curves.}

\begin{figure}

\centering
\begin{subfigure}[t]{0.3\columnwidth}
  \centering
  \includegraphics[width=\columnwidth,trim=0.in 0.in 0.in 0.in,clip]{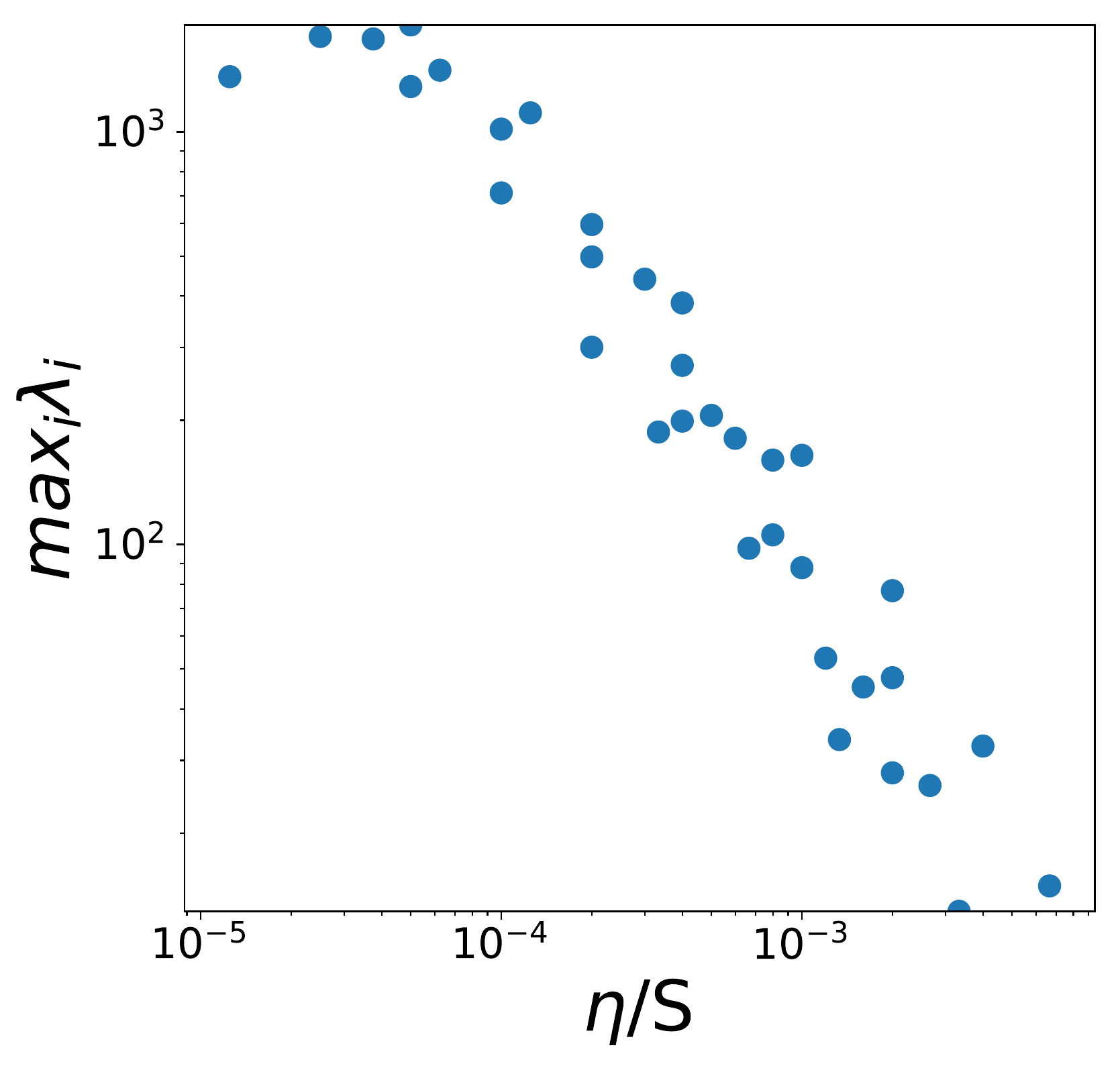}
  \caption{Largest eigenvalue. }
  \label{fig:1_deeprelu_LRBS_correlation_HE}
\end{subfigure}
~
\begin{subfigure}[t]{0.3\columnwidth}
  \centering
  \includegraphics[width=\columnwidth,trim=0.in 0.in 0.in 0.in,clip]{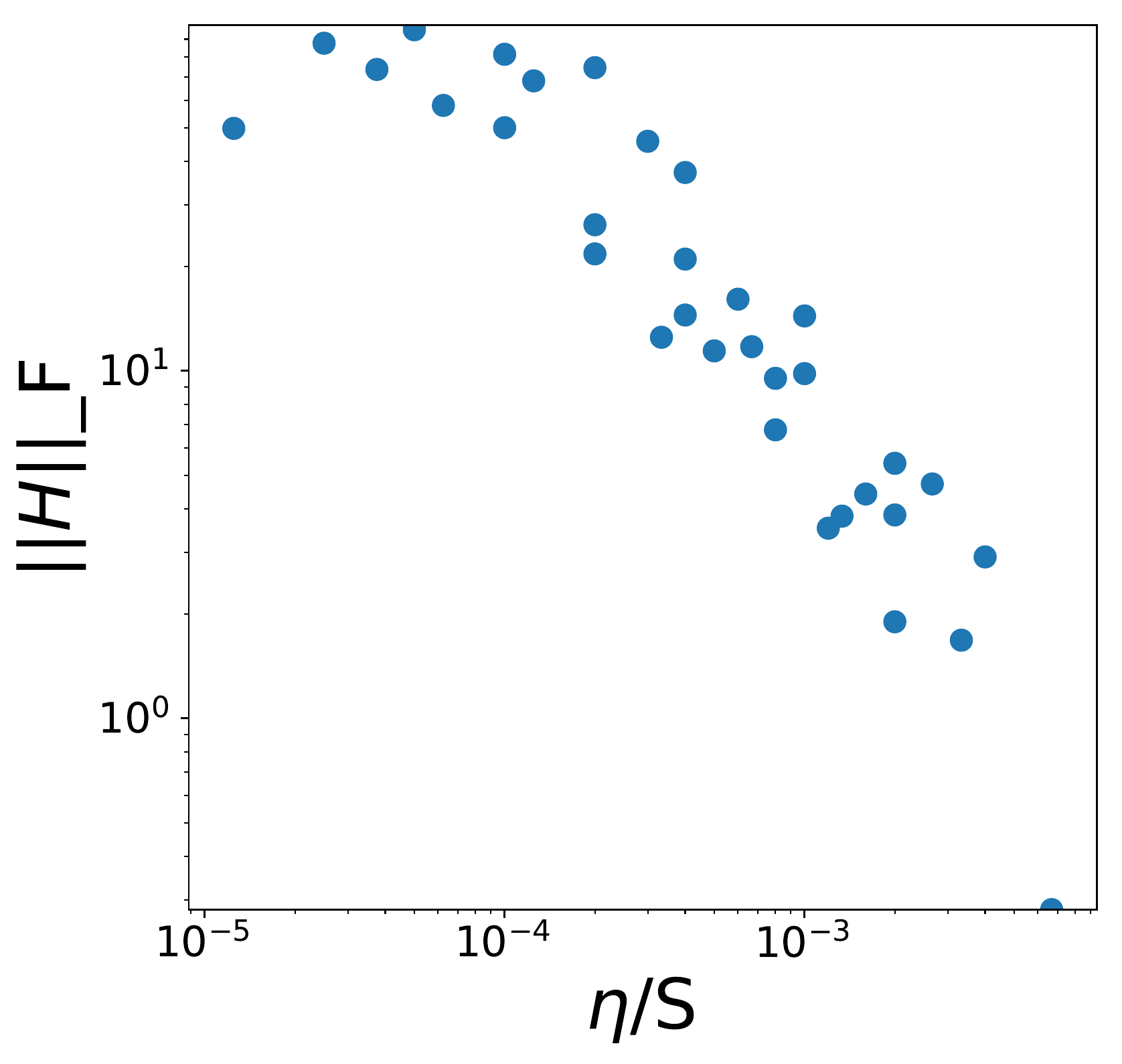}
  \caption{Frobenius norm. }
  \label{fig:1_deeprelu_LRBS_correlation_HN}
\end{subfigure}
~
\begin{subfigure}[t]{0.3\columnwidth}
  \centering
 \includegraphics[width=\columnwidth,trim=0.in 0.in 0.in 0.in,clip]{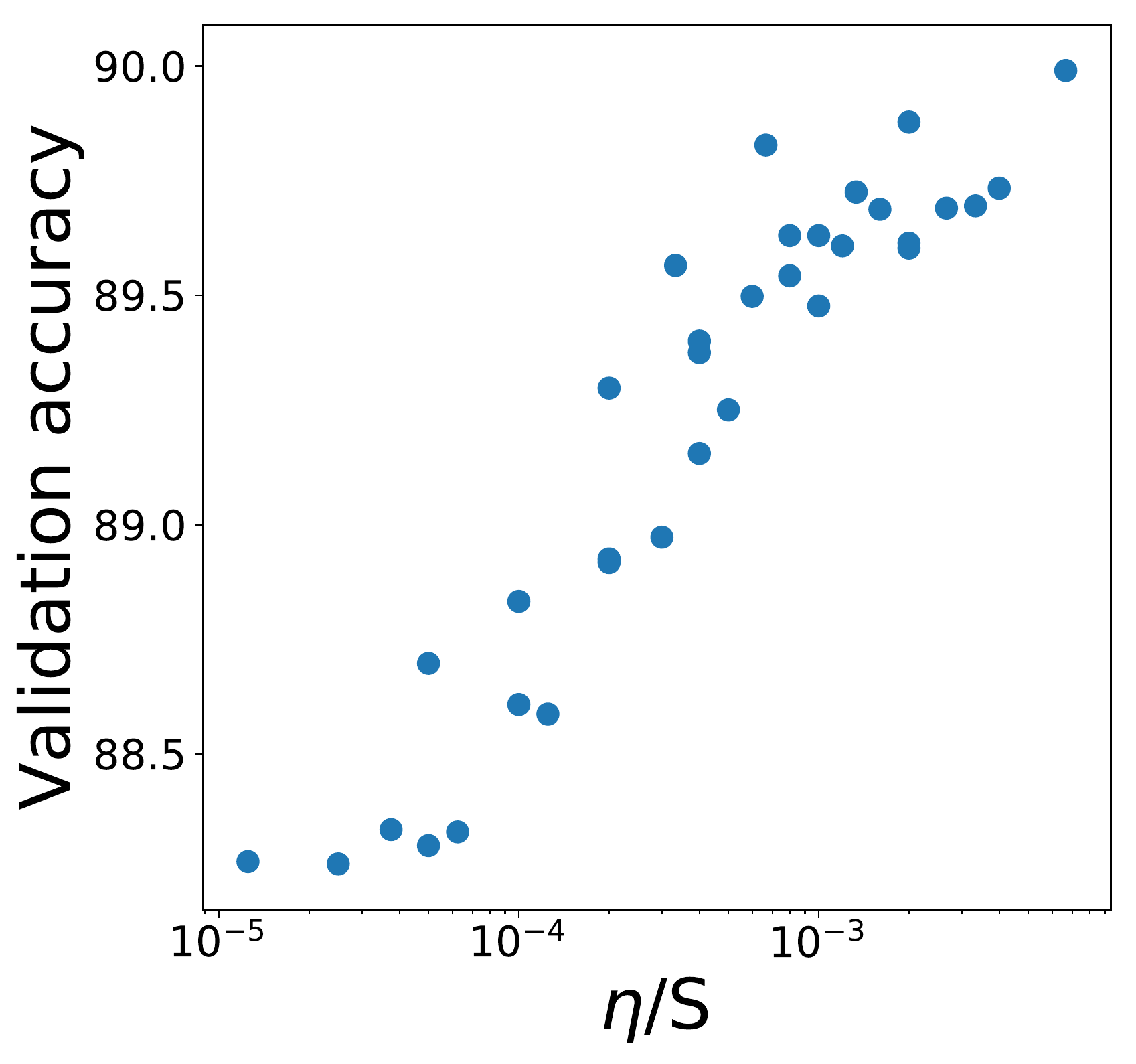}
 \caption{Validation accuracy.}
 \label{fig:deeprelu_LRBS_correlation}
\end{subfigure}
\caption{Ratio of learning rate to batch size, $\eta/S$, for a grid of $\eta$, $S$ for 4 layer ReLU MLP on FashionMNIST. Higher $\eta/S$ correlates with lower Hessian maximum eigenvalue, lower Hessian Frobenius norm, i.e. wider minima, and better generalization. The validation accuracy is similar for different batch sizes, and different learning rates, so long as the ratio is constant.}

\end{figure}

\subsection{Geometry and generalization depend on LR/BS}
In this section we investigate experimentally the {impact of learning rate to batch size ratio} on the geometry of the region that SGD ends in. 
We trained a series of 4-layer batch-normalized ReLU MLPs  on Fashion-MNIST \cite{2017arXiv170807747X} with different $\eta, S$\footnote{{Each experiment {was} run for $300$ epochs. Models reaching an accuracy of approximately $100\%$ on the training set were selected.}}. 
To 
{access}
the {loss} curvature at the end of training, we computed the largest eigenvalue and we approximated the Frobenius norm of the Hessian\footnote{{The largest eigenvalue and the Frobenius norm of the Hessian are used in place of the trace of the Hessian, because calculating multiple eigenvalues to directly approximate the trace is too computationally intensive.}} (higher values imply a  sharper minimum) using the finite difference {method}~\cite{wu2017towards}.
Fig.~\ref{fig:1_deeprelu_LRBS_correlation_HE} and Fig.~\ref{fig:1_deeprelu_LRBS_correlation_HN} show {the values of these quantities}
for 
minima obtained by SGD for different 
$\frac{\eta}{S}$, {with }$\eta\in[5e-3, 1e-1]$ and $S\in[25, 1000]$.  As $\frac{\eta}{S}$ grows, 
the norm of the Hessian at the minimum 
decreases, suggesting that higher {values of} $\frac{\eta}{S}$ push the optimization towards flatter regions. 
Figure~\ref{fig:deeprelu_LRBS_correlation} shows the results from exploring  the impact of $\frac{\eta}{S}$ on the final validation performance, which confirms that better generalization correlates with higher {values of $\frac{\eta}{S}$}. Taken together, Fig.~\ref{fig:1_deeprelu_LRBS_correlation_HE}, Fig.~\ref{fig:1_deeprelu_LRBS_correlation_HN} and Fig.~~\ref{fig:deeprelu_LRBS_correlation} imply {that} as $\frac{\eta}{S}$ increases, SGD finds wider regions which correlate well with better generalization\footnote{{assuming}
the network has enough capacity}. 

\begin{figure}

 \centering
\begin{subfigure}[t]{0.45\columnwidth}
  \centering
  \includegraphics[width=\columnwidth ,trim=0.in 0.in 0.in 0.in,clip]{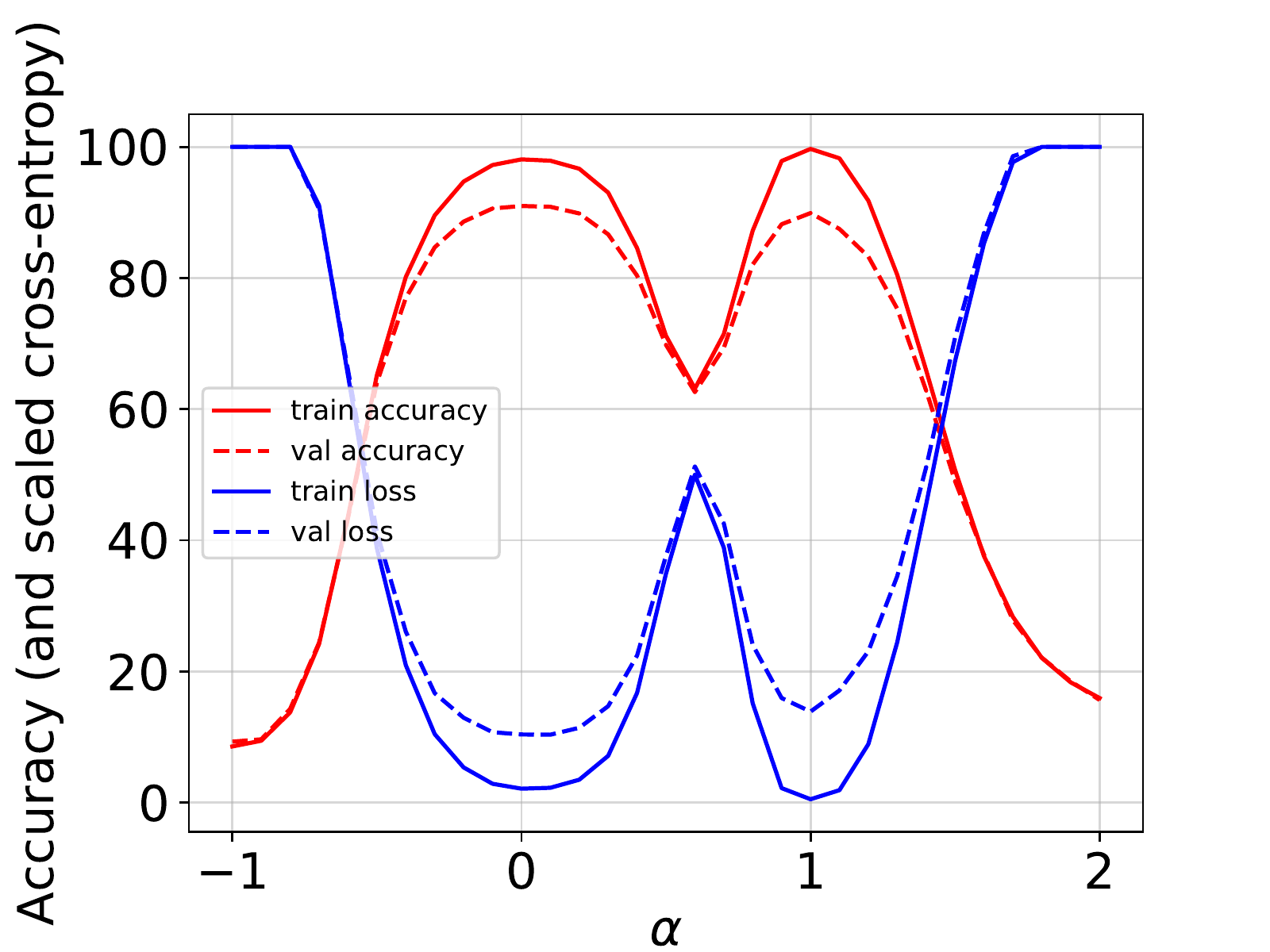}
  \caption{{ $\left[ \frac{\eta=0.1}{S=128}, \frac{\eta=0.1}{S=1024}\right]$}}
\end{subfigure}
\begin{subfigure}[t]{0.45\columnwidth}
  \centering
  \includegraphics[width=\columnwidth ,trim=0.in 0.in 0.in 0.in,clip]{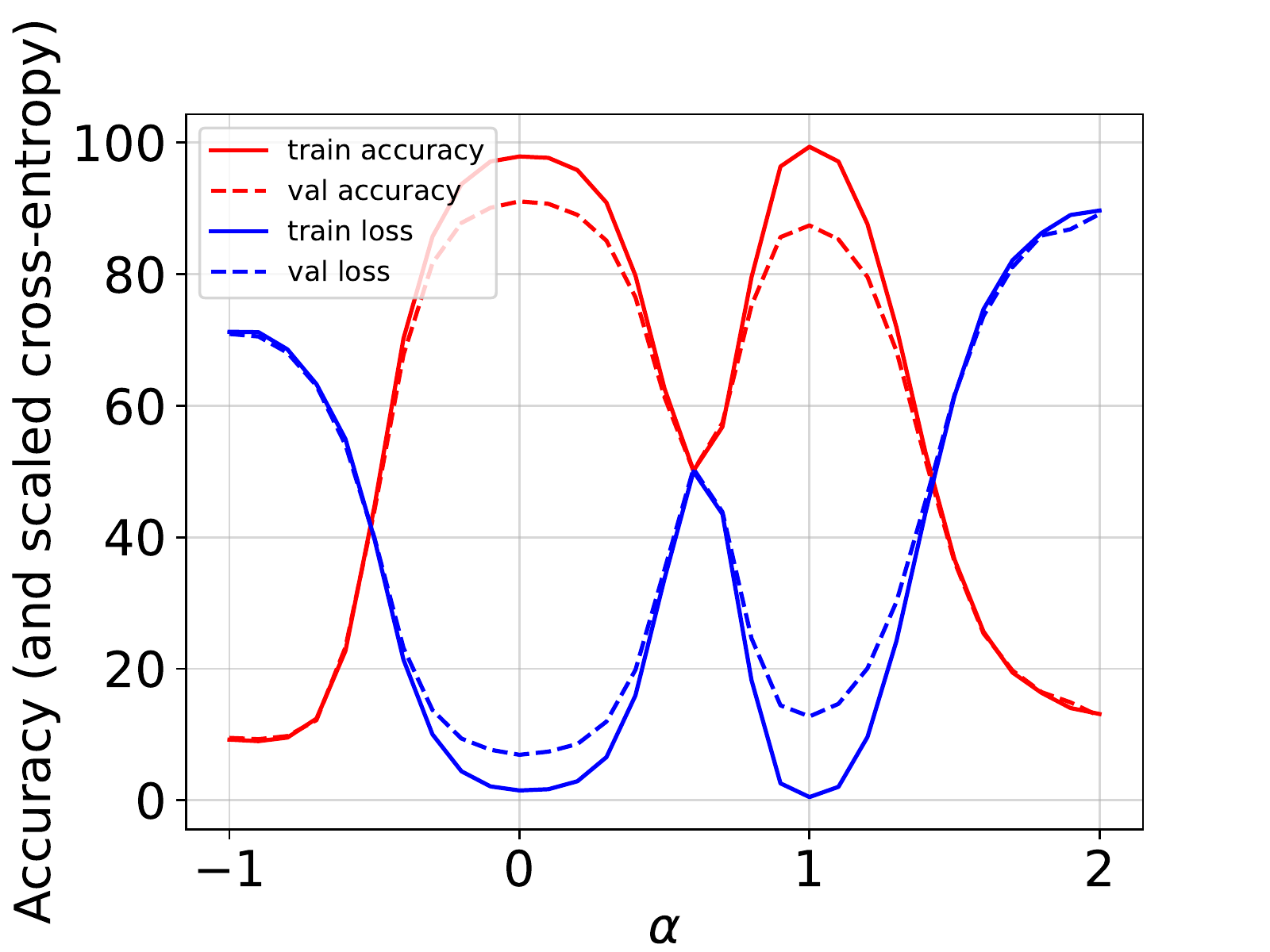}
  \caption{{ $\left[\frac{\eta=0.1}{S=128},\frac{\eta=0.01}{S=128}\right]$}}

\end{subfigure}
\\
\begin{subfigure}[t]{0.45\columnwidth}
  \centering
  \includegraphics[width=\columnwidth ,trim=0.in 0.in 0.in 0.in,clip]{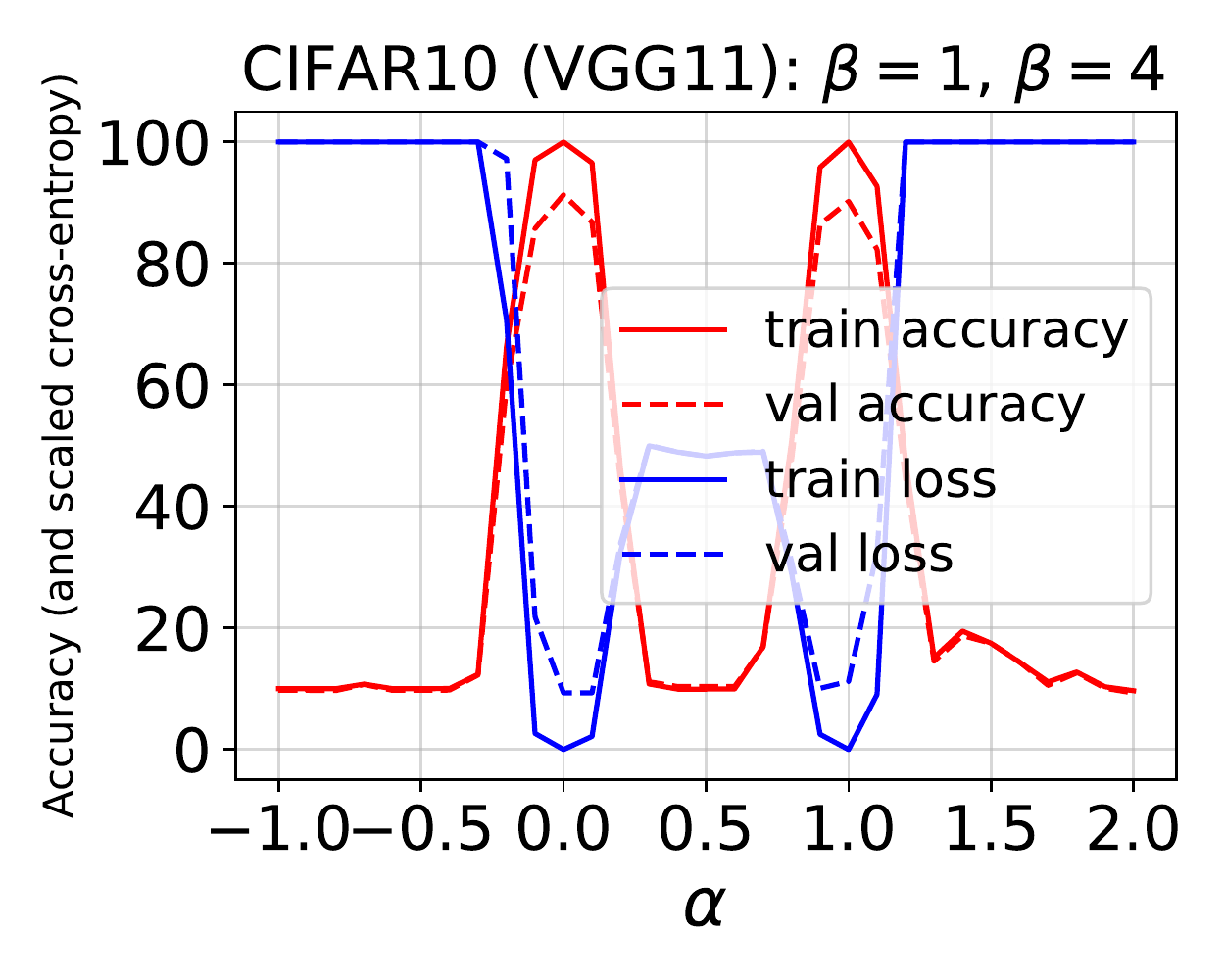}
  \caption{{ $\left[\frac{\eta=0.1 }{S=50 }, \frac{\eta=0.1 \times 4}{S=50 \times 4}\right]$}}
\end{subfigure}
\begin{subfigure}[t]{0.45\columnwidth}
  \centering
  \includegraphics[width=\columnwidth ,trim=0.in 0.in 0.in 0.in,clip]{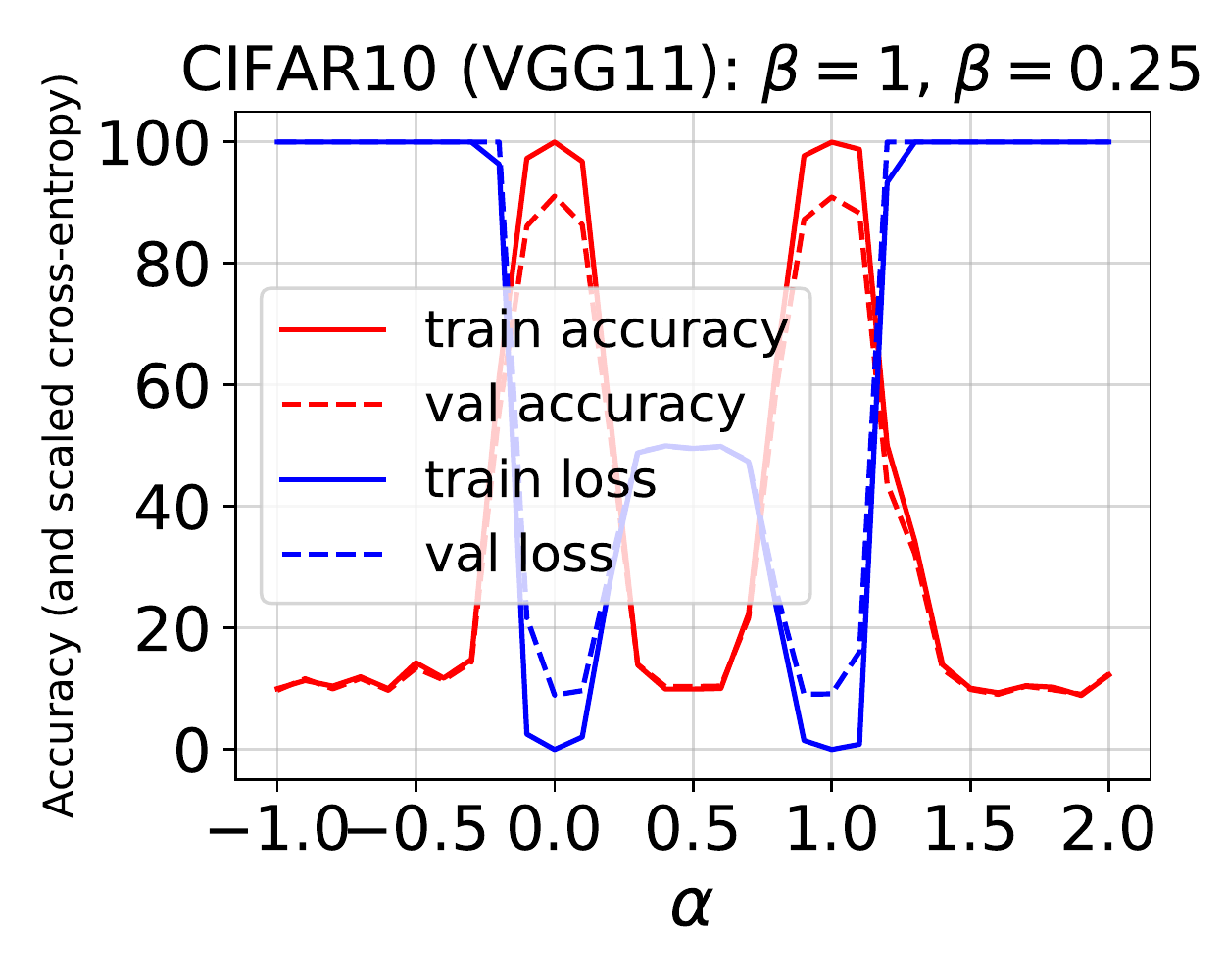}
  \caption{{ $\left[\frac{\eta=0.1 }{S=50 },\frac{\eta=0.1 \times 0.25}{S=50 \times 0.25}\right]$ }}
\end{subfigure}
\caption{Interpolations between models with $\alpha$ interpolation coefficient. At $\alpha=0$ there is one trained model (1st element of subcaption), at $\alpha=1$ there is another (2nd element of subcaption). (a), (b): Resnet56 with different ratio $\frac{\eta}{S}$.  (c), (d): VGG11 with the same ratio, but different $\eta, S$. Higher ratios give wider minima (a,b) as seen by the great width of the basin around $\alpha=0$, whilst the same ratio gives the same width minima (c,d), despite differences in batch size and learning rate.}
\label{fig:same_noise_diff_bp}

\end{figure}

In Fig.~\ref{fig:same_noise_diff_bp} we qualitatively illustrate the behavior of SGD with different 
$\frac{\eta}{S}$. We follow~\cite{goodfellow2014qualitatively} {by investigating the loss
on the line interpolating between the parameters of two models} with interpolation coefficicent $\alpha$.
In Fig.~\ref{fig:same_noise_diff_bp}(a,b) we consider Resnet56 models on CIFAR10 for different $\frac{\eta}{S}$. We see sharper regions on the right of each, for the lower $\frac{\eta}{S}$.
In Fig.~\ref{fig:same_noise_diff_bp}(c,d) we consider VGG-11 models on CIFAR10 for the same ratio, but different $\beta$, where $\frac{\eta=0.1 \times \beta}{S=50 \times \beta}$. We see the same sharpness for the same ratio.
Experiments were repeated several times with different random initializations and qualitatively similar plots were achieved.

\begin{figure}

\centering
\begin{subfigure}[b]{0.47\columnwidth}
  \centering
  \includegraphics[width=1\columnwidth]{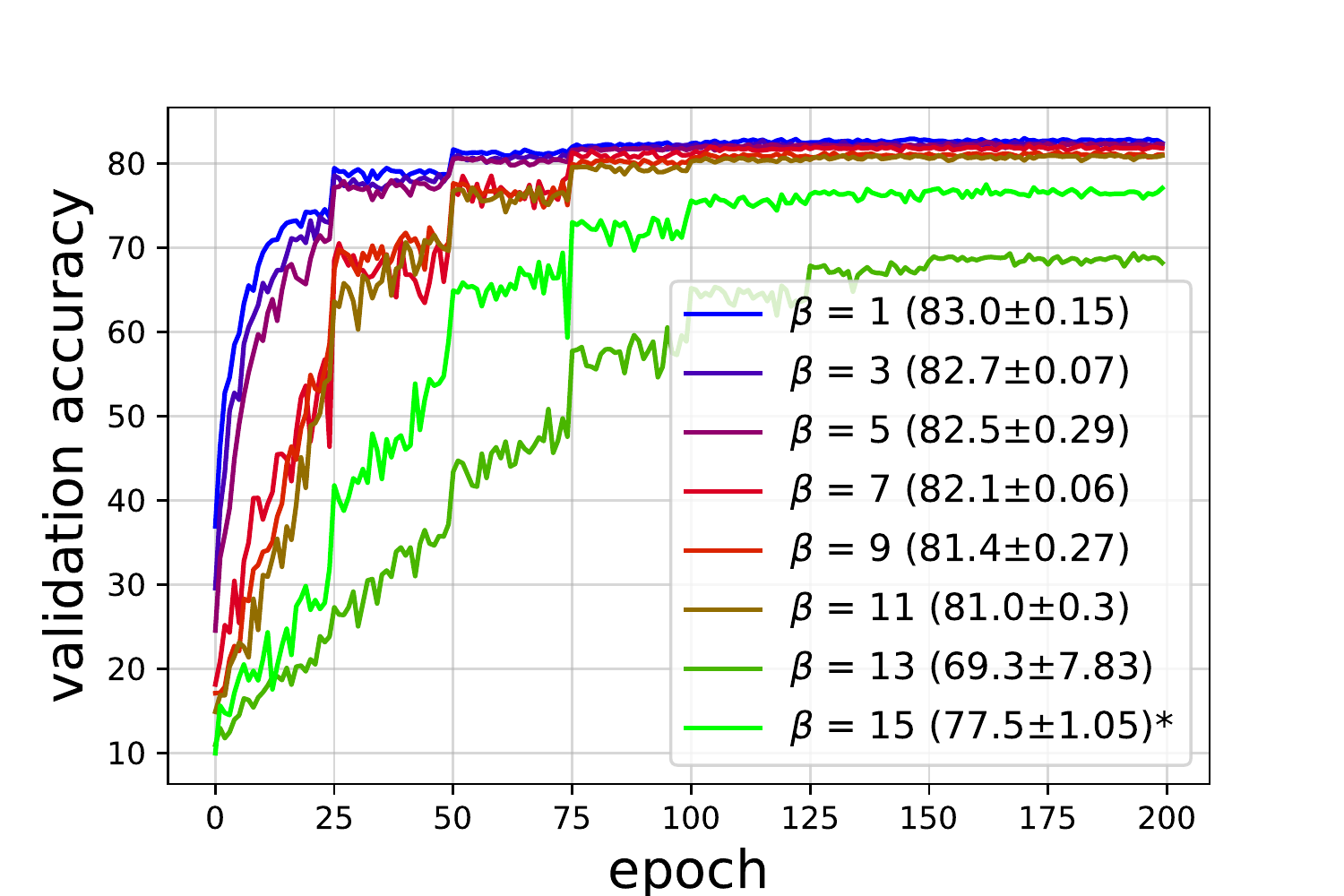}
  \caption{Train dataset size 12000}
\end{subfigure}
\begin{subfigure}[b]{0.47\columnwidth}
  \centering
  \includegraphics[width=1.\columnwidth]{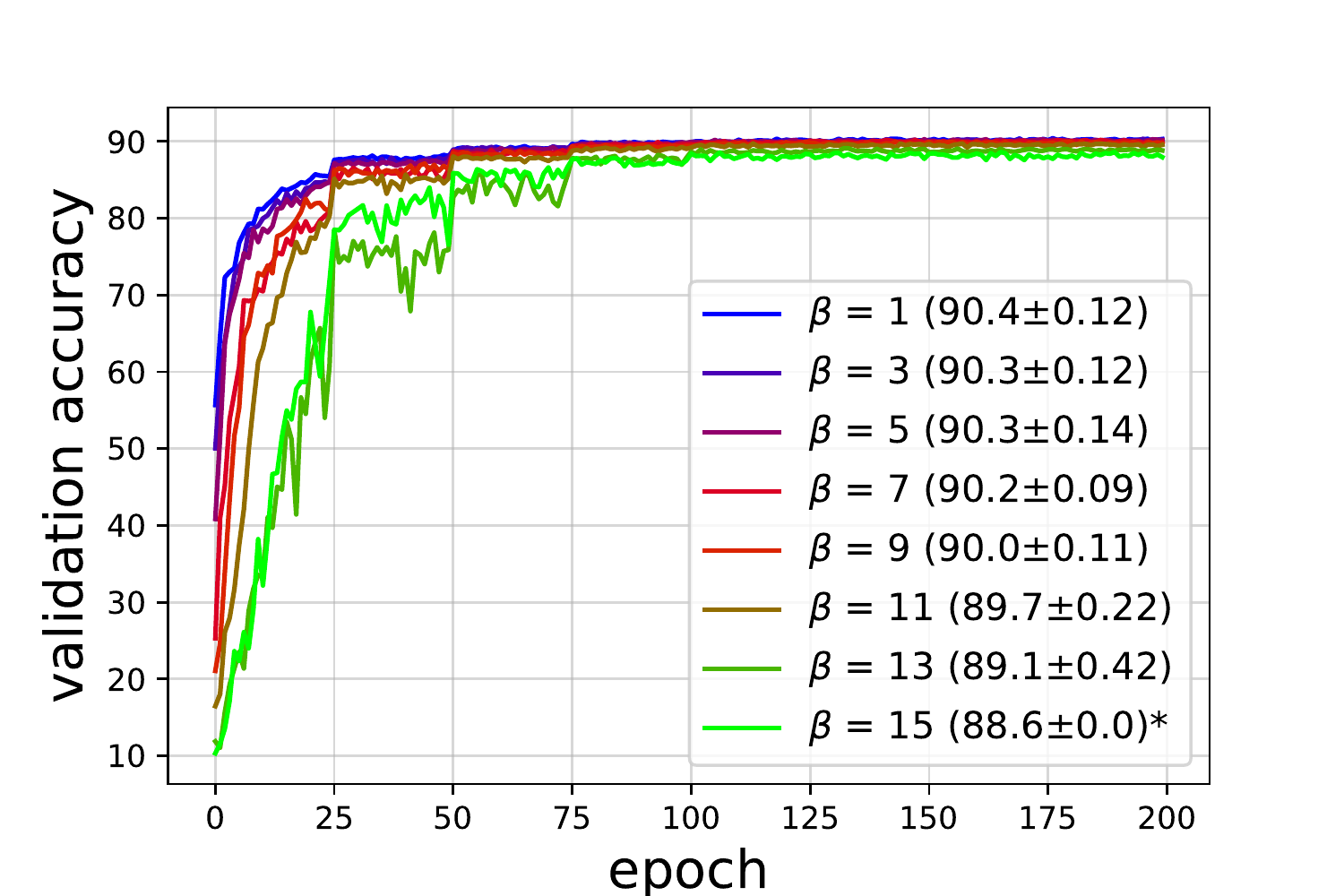}
  \caption{Train dataset size 45000}
\end{subfigure}
\caption{Validation accuracy for different dataset sizes and different $\beta$ values for fixed ratio $ \frac{\beta \times (\eta=0.1) }{ \beta \times (S=50)}$. The curves diverging from the blue shows the approximation of the SDE discretized to SGD breaking down for large $\beta$, which is magnified for smaller dataset size.
}
\label{fig:BP_vgg11_cifar10}

\end{figure}

\subsection{Cyclic schedules}

It has been observed that a cyclic learning rate (CLR) schedule leads to better generalization~\cite{2015arXiv150601186S}. We have demonstrated that one can exchange a cyclic learning rate schedule (CLR) with a cyclic batch size (CBS) and approximately preserve the practical benefit of CLR. This exchangeability shows that the generalization benefit of CLR must come from the varying noise level, rather than just from cycling the learning rate. To explore why this helps generalization, we run VGG-11 on CIFAR10 using 5 training schedules: we compared a discrete cyclic learning rate, a discrete cyclic batch size, a triangular cyclic learning rate and a baseline constant learning rate.
We track throughout training the Frobenius norm of the Hessian (divided by number of parameters, $D$), the largest eigenvalue of the Hessian, and the training loss. For each schedule we optimize both $\eta$ in the range [$1e-3$, $5e-2$] and the cycle length from $\{5, 10, 15\}$ on a validation set. In all cyclical schedules the maximum value (of $\eta$ or S) is $5\times$ larger than the minimum value. Sharpness is measured at the best validation score.

The results are shown in Table~\ref{tab:cyc}.
First we note that all cyclic schedules lead to wider bowls (both in terms of Frobenius norm and the largest eigenvalue) and higher loss values than the baseline. 
We note the discrete $S$ schedule leads to much wider bowls for a similar value of the loss. 
We also note that the discrete schedules varying either $S$ or $\eta$ performs similarly, or slightly better than triangular CLR schedule.
The results suggest that by by being exposed to higher noise levels, cyclical schemes reach wider endpoints at higher loss than constant learning rate schemes with the same final noise level. 
We leave the exploration of the implications for cyclic schedules and a more thorough comparison with other noise schedules for future work.

\begin{table}
\centering
\small
\resizebox{\columnwidth}{!}{%

\begin{tabular}{llllll}
\toprule
{} & $max\_i \lambda\_i$ & $||H|| / D$ &               Loss &               Test acc. &              Valid acc. \\
\midrule
Discrete $\eta$ &           $10.96$ &      $0.42$ &  $0.054 \pm 0.000$ &  $90.23\% \pm 0.02\% $ &  $90.52\% \pm 0.42\% $ \\
Discrete S      &            $9.04$ &      $0.52$ &  $0.057 \pm 0.004$ &  $90.24\% \pm 0.03\% $ &  $90.20\% \pm 0.08\% $ \\
Triangle $\eta$ &            $9.79$ &      $0.37$ &  $0.067 \pm 0.001$ &  $89.94\% \pm 0.02\% $ &  $89.86\% \pm 0.11\% $ \\
\bottomrule
Constant        &           $19.46$ &      $1.50$ &  $0.056 \pm 0.001$ &  $88.17\% \pm 0.30\% $ &  $88.33\% \pm 0.01\% $ \\
\bottomrule
\end{tabular}

}
\caption{Comparison between different cyclical training schedules (cycle length and learning rate are optimized using a grid search). }
\label{tab:cyc}
\end{table}

 \begin{figure}
\centering
\begin{subfigure}[b]{\textwidth}
\centering
\includegraphics[width=0.45\columnwidth,trim=0.in 0.in 0.in 0.in,clip]{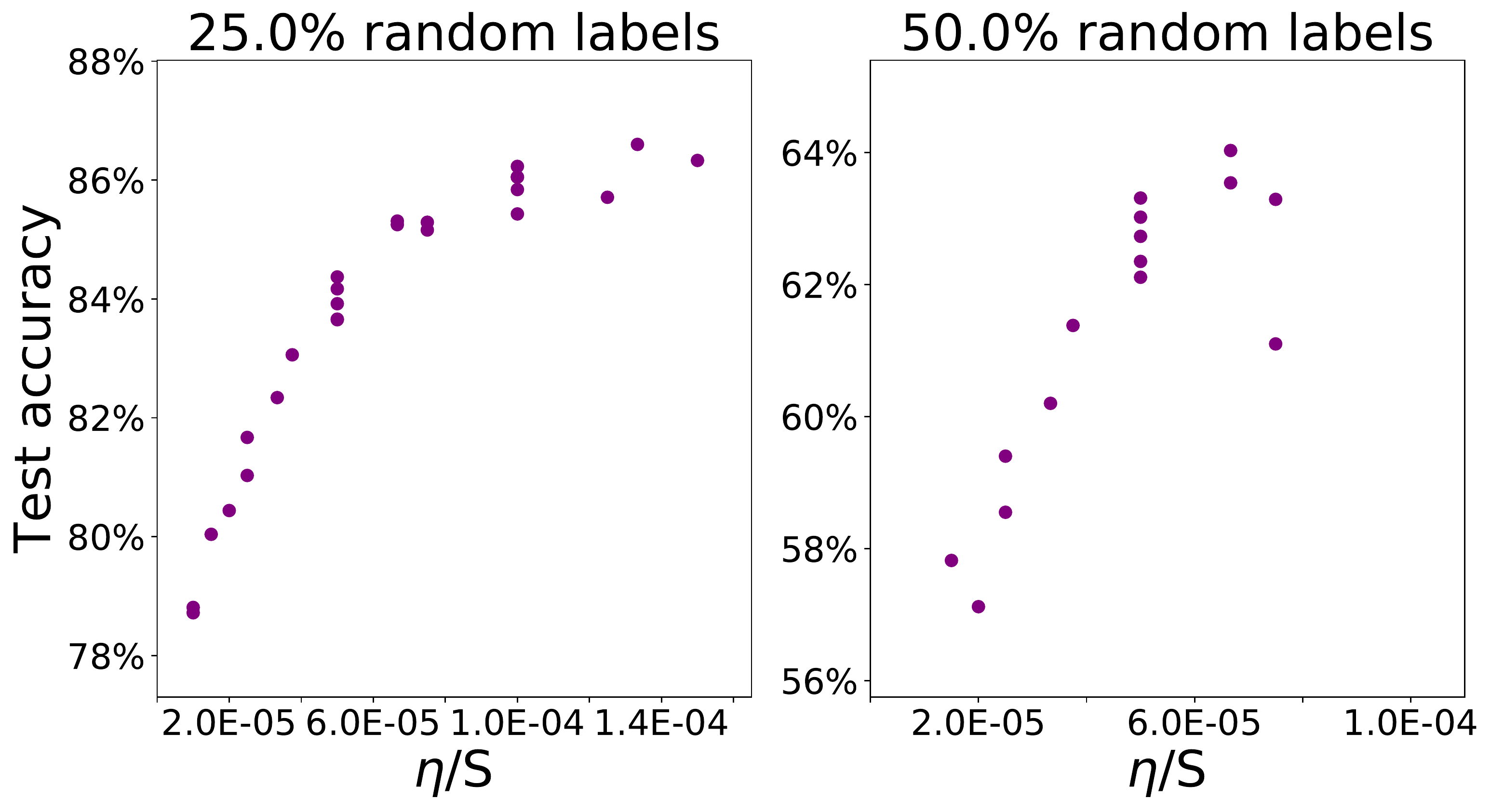}
\includegraphics[width=0.45\columnwidth,trim=0.in 0.in 0.in 0.in,clip]{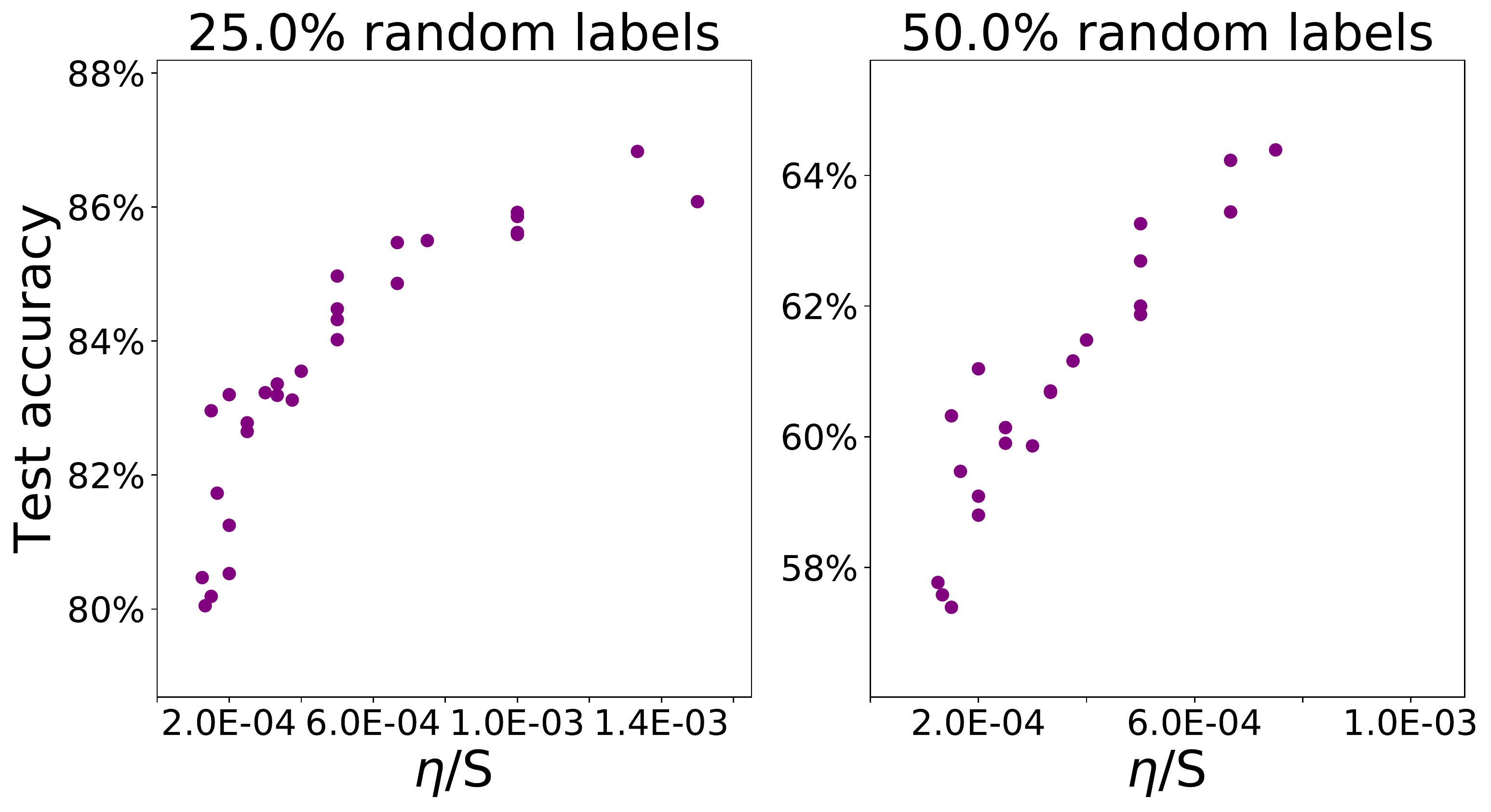}
\end{subfigure}
 \caption{ Impact of $\frac{\eta}{S}$ on memorization of MNIST when $25\%$ and $50\%$ of labels in the training set are replaced with random labels,
using no momentum (on the right) or a momentum with parameter $0.9$ (on the left).
We observe that high $\frac{\eta}{S}$ leads to better generalization under full memorization of the training set. 
}
\label{fig:mem_vs_noise}
 \end{figure}

\subsection{Impact of SGD on memorization}
To generalize well, a model must identify the underlying pattern in the data instead of simply perfectly memorizing each training example. An empirical approach to test for memorization is to analyze how good a DNN can fit a training set when the true labels are partly replaced by random labels~\cite{zhang2016understanding, pmlr-v70-arpit17a}. To better characterize the practical benefit of ending in a wide bowl, we look at memorization of the training set under varying levels of learning rate to batch size ratio.
The experiments described in this section
highlight that SGD with a sufficient amount of noise improves generalization after memorizing the training set.

Experiments are performed on the MNIST dataset with an MLP similar to the one used by \cite{pmlr-v70-arpit17a}, but with $256$ hidden units. We train the MLP 
with different amounts of random labels in the training set ($25\%$ and $50\%$). For each level of label noise, we evaluate the impact of $\frac{\eta}{S}$ 
on the generalization performance.
Specifically, we run experiments with $\frac{\eta}{S}$ taking values in a grid with batch size in range $[25, 1000]$, learning rate in range $[0.005, 0.1]$, and momentum in $\{0.0, 0.9\}$. Models are trained for $1000$ epochs. Fig.~\ref{fig:mem_vs_noise} reports the MLPs performances on both the noisy training set and the validation set after memorizing the training set 
(defined here as achieving $\geq 99.9\%$ accuracy on random labels).
The results show that larger noise in SGD (regardless if 
induced by
using a smaller batch size or a larger learning rate) leads to solutions which generalize better 
after having memorized the training set. Additionally we observe as in previous Sections a strong correlation of the Hessian norm with $\frac{\eta}{S}$ ($-0.58$ with p-value $\leq 0.001$).
We highlight that SGD with low noise $n = \frac{\eta}{S}$ steers the endpoint of optimization towards a minimum with low generalization ability.

\subsection{Breakdown of $\eta/S$ scaling}

We expect {discretization} errors to become important when the learning rate gets large, we expect our central limit theorem to break down for large batch size and smaller dataset size. 

We show this experimentally in Fig.~\ref{fig:BP_vgg11_cifar10}, where similar learning dynamics and final performance can be observed when simultaneously multiplying the learning rate and batch size by a factor $\beta$ up to a certain limit\footnote{Experiments
{are repeated} 5 times with different {random} seeds. The graphs denote the mean validation accuracies and the numbers in the brackets denote the mean and standard deviation of the maximum validation accuracy across different runs. The * denotes at least one seed diverged.}. This is done for a smaller training set size in Fig.~\ref{fig:BP_vgg11_cifar10} (a) than in (b). 
The curves don't match when $\beta$ gets too large as expected from our approximations.

\section{Related work}\label{sec:related}

The analysis of SGD as an SDE is well established in the stochastic approximation literature, see e.g. \cite{ljung1992} and \cite{kushner2013stochastic}.
It was shown by \cite{pmlr-v70-li17f} that SGD can be approximated by an SDE in an order-one weak approximation. However, batch size does not enter their analysis. 
In contrast, our analysis makes the role of batch size evident and shows the dynamics are set by the ratio of learning rate to batch size.
\cite{2017arXiv170507562J} reproduce the SDE result of \cite{pmlr-v70-li17f} and further show that the covariance matrix of the minibatch-gradient scales inversely with the batch size\footnote{This holds approximately, in the limit of small batch size compared to training set size.} and proportionally to the sample covariance matrix over all examples in the training set.
\cite{Mandt2017StochasticGD} approximate SGD by a different SDE and show that SGD can be used as an approximate Bayesian posterior inference algorithm. In contrast, we show the ratio of learning rate over batch influences the width of the minima found by SGD. We then explore each of these experimentally linking also to generalization.

Many works have used stochastic gradients to sample from a posterior, see e.g. \cite{Welling2011}, using a decreasing learning rate to correctly sample from the actual posterior. In contrast, we consider SGD with a fixed learning rate and our focus is not on applying SGD to sample from the actual posterior.

Our work is closely related to the ongoing discussion about how batch size affects sharpness and generalization.
Our work extends this by investigating the impact of both batch size and learning rate on sharpness and generalization. 
\cite{2016arXiv160904836S} showed empirically that SGD ends up in a sharp minimum when using a large batch size. \cite{2017arXiv170508741H} rescale the learning rate with the square root of the batch size, and train for more epochs to reach the same generalization with a large batch size.
The empirical analysis of \cite{2017arXiv170602677G} demonstrated that rescaling the learning rate linearly with batch size can result in same generalization. Our 
work theoretically explains this empirical finding, and extends the experimental results on this.

Anisotropic noise in SGD was studied in \cite{2018arXiv180300195Z}. It was found that the gradient covariance matrix is approximately the same as the Hessian, late on in training. 
In the work of \cite{2017arXiv170604454S}, the Hessian is also related to the gradient covariance matrix, and both are found to be highly anisotropic. In contrast, our focus is on the importance of the scale of the noise, set by the learning rate to batch size ratio.

Concurrent with this work, \cite{smith2017understanding} derive an analytical expression for the stochastic noise scale and -- based on the trade-off between depth and width in the Bayesian evidence -- find an optimal noise scale for optimizing the test accuracy. 
\cite{chaudhari2017stochastic} explored the stationary non-equilibrium solution for the SDE for non-isotropic gradient noise.

{In contrast to these concurrent works, our emphasis is on how the learning rate to batch size ratio relates to the width of the minima sampled by SGD. We show theoretically that different SGD processes with the same ratio are different discretizations of the same underlying SDE and hence follow the same dynamics. Further their learning curves will match under simultaneous rescaling of the learning rate and batch size when plotted on an epoch time axis. We also show that at the end of training, the learning rate to batch size ratio affects the width of the regions that SGD ends in, and empirically verify that the width of the endpoint region correlates with the learning rate to batch size ratio in practice. }

\section{Conclusion}

{
In this paper we investigated a relation between learning rate, batch size and the properties of the final minima. By approximating SGD as an SDE, we found that the learning rate to batch size ratio controls the dynamics by scaling the stochastic noise. Furthermore, under the discussed assumption on the relation of covariance of gradients and the Hessian, the ratio is a key determinant of width of the minima found by SGD. The learning rate, batch size and the covariance of gradients, in its link to the Hessian, are three factors influencing the final minima.}

{We experimentally explored this relation using a range of DNN models and datasets, finding approximate invariance under rescaling of learning rate and batch size, and that the ratio of learning rate to batch size correlates with width and generalization with a higher ratio leading to wider minima and better generalization. Finally, our experiments suggest schedules with a changing batch size during training are a viable alternative to a changing learning rate. }

{\small 
 {\bf Acknowledgements}
 We thank Agnieszka Pocha, Jason Jo, Nicolas Le Roux, Mike Rabbat, Leon Bottou, and James Griffin for discussions. We thank NSERC, Canada Research Chairs, IVADO and CIFAR for funding. SJ was in part supported by Grant No.~DI 2014/016644 from Ministry of Science and Higher Education, Poland and ETIUDA stipend No.~2017/24/T/ST6/00487 from National Science Centre, Poland. We acknowledge the computing resources provided by ComputeCanada and CalculQuebec. This project has received funding from the European Union’s Horizon 2020 research and innovation programme under grant agreement No 732204 (Bonseyes). This work is supported by the Swiss State Secretariat for Education‚ Research and Innovation (SERI) under contract number 16.0159. The opinions expressed and arguments employed herein do not necessarily reflect the official views of these funding bodies.

\bibliographystyle{arxiv}
\bibliography{references}

\begin{thebibliography}{36}
\providecommand{\natexlab}[1]{#1}
\providecommand{\url}[1]{\texttt{#1}}
\expandafter\ifx\csname urlstyle\endcsname\relax
  \providecommand{\doi}[1]{doi: #1}\else
  \providecommand{\doi}{doi: \begingroup \urlstyle{rm}\Url}\fi

\bibitem[Advani \& Saxe(2017)Advani and Saxe]{advani2017high}
M.~S. Advani and A.~M. Saxe.
\newblock High-dimensional dynamics of generalization error in neural networks.
\newblock \emph{arXiv preprint arXiv:1710.03667}, 2017.

\bibitem[Amari(1998)]{Amari:1998:NGW:287476.287477}
Shun-Ichi Amari.
\newblock Natural gradient works efficiently in learning.
\newblock \emph{Neural Comput.}, 10\penalty0 (2):\penalty0 251--276, February
  1998.
\newblock ISSN 0899-7667.
\newblock \doi{10.1162/089976698300017746}.
\newblock URL \url{http://dx.doi.org/10.1162/089976698300017746}.

\bibitem[Arpit \& et~al.(2017)Arpit and et~al.]{pmlr-v70-arpit17a}
D.~Arpit and et~al.
\newblock A closer look at memorization in deep networks.
\newblock In \emph{ICML}, 2017.

\bibitem[Bottou(1998)]{bottou1998online}
L.~Bottou.
\newblock Online learning and stochastic approximations.
\newblock \emph{On-line learning in neural networks}, 17\penalty0 (9):\penalty0
  142, 1998.

\bibitem[Chaudhari \& Soatto(2017)Chaudhari and
  Soatto]{chaudhari2017stochastic}
P.~Chaudhari and S.~Soatto.
\newblock Stochastic gradient descent performs variational inference, converges
  to limit cycles for deep networks.
\newblock \emph{arXiv:1710.11029}, 2017.

\bibitem[{Dinh} et~al.(2017){Dinh}, {Pascanu}, {Bengio}, and
  {Bengio}]{2017arXiv170304933D}
L.~{Dinh}, R.~{Pascanu}, S.~{Bengio}, and Y.~{Bengio}.
\newblock {Sharp Minima Can Generalize For Deep Nets}.
\newblock \emph{ArXiv e-prints}, 2017.

\bibitem[Gardiner()]{gardiner2010stochastic}
C.~Gardiner.
\newblock \emph{Stochastic Methods: A Handbook for the Natural and Social
  Sciences}.
\newblock Springer Series in Synergetics.
\newblock ISBN 9783642089626.

\bibitem[Goodfellow et~al.(2014)Goodfellow, Vinyals, and
  Saxe]{goodfellow2014qualitatively}
I.~J. Goodfellow, O.~Vinyals, and A.~M. Saxe.
\newblock Qualitatively characterizing neural network optimization problems.
\newblock \emph{arXiv preprint arXiv:1412.6544}, 2014.

\bibitem[{Goyal} \& et~al.(2017){Goyal} and et~al.]{2017arXiv170602677G}
P.~{Goyal} and et~al.
\newblock {Accurate, Large Minibatch SGD: Training ImageNet in 1 Hour}.
\newblock \emph{ArXiv e-prints}, 2017.

\bibitem[Heskes \& Kappen(1993)Heskes and Kappen]{HESKES1993199}
T.~M. Heskes and B.~Kappen.
\newblock On-line learning processes in artificial neural networks.
\newblock volume~51 of \emph{North-Holland Mathematical Library}, pp.\  199 --
  233. Elsevier, 1993.
\newblock \doi{https://doi.org/10.1016/S0924-6509(08)70038-2}.
\newblock URL
  \url{http://www.sciencedirect.com/science/article/pii/S0924650908700382}.

\bibitem[Hochreiter \& Schmidhuber(1997)Hochreiter and
  Schmidhuber]{hochreiter1997flat}
S.~Hochreiter and J.~Schmidhuber.
\newblock Flat minima.
\newblock \emph{Neural Computation}, 9\penalty0 (1):\penalty0 1--42, 1997.

\bibitem[{Hoffer} \& et~al.(){Hoffer} and et~al.]{2017arXiv170508741H}
E.~{Hoffer} and et~al.
\newblock {Train longer, generalize better: closing the generalization gap in
  large batch training of neural networks}.
\newblock \emph{ArXiv e-prints, arxiv:1705.08741}.

\bibitem[{Junchi Li} \& et~al.(2017){Junchi Li} and
  et~al.]{2017arXiv170507562J}
C.~{Junchi Li} and et~al.
\newblock {Batch Size Matters: A Diffusion Approximation Framework on Nonconvex
  Stochastic Gradient Descent}.
\newblock \emph{ArXiv e-prints}, 2017.

\bibitem[Kass \& Raftery(1995)Kass and
  Raftery]{doi:10.1080/01621459.1995.10476572}
R.~E. Kass and A.~E. Raftery.
\newblock Bayes factors.
\newblock \emph{Journal of the American Statistical Association}, 90\penalty0
  (430):\penalty0 773--795, 1995.
\newblock \doi{10.1080/01621459.1995.10476572}.
\newblock URL
  \url{http://amstat.tandfonline.com/doi/abs/10.1080/01621459.1995.10476572}.

\bibitem[Kloeden \& Platen(1992)Kloeden and Platen]{kloeden}
Peter~E. Kloeden and Eckhard Platen.
\newblock \emph{{Numerical Solution of Stochastic Differential Equations}}.
\newblock Springer, 1992.
\newblock ISBN 978-3-662-12616-5.

\bibitem[Kushner \& Yin()Kushner and Yin]{kushner2013stochastic}
H.~Kushner and G.G. Yin.
\newblock \emph{Stochastic Approximation and Recursive Algorithms and
  Applications}.
\newblock Stochastic Modelling and Applied Probability.
\newblock ISBN 9781489926968.

\bibitem[Li et~al.(2017)Li, Tai, and E.]{pmlr-v70-li17f}
Q.~Li, C.~Tai, and Weinan E.
\newblock Stochastic modified equations and adaptive stochastic gradient
  algorithms.
\newblock In \emph{Proceedings of the 34th ICML}, 2017.

\bibitem[Ljung et~al.(1992)Ljung, Pflug, and Walk]{ljung1992}
L.~Ljung, G.~Pflug, and H.~Walk.
\newblock 1992.

\bibitem[MacKay(1992)]{6796869}
D.~J.~C. MacKay.
\newblock A practical bayesian framework for backpropagation networks.
\newblock \emph{Neural Computation}, 4\penalty0 (3):\penalty0 448--472, 1992.
\newblock ISSN 0899-7667.
\newblock \doi{10.1162/neco.1992.4.3.448}.

\bibitem[Mandt et~al.(2017)Mandt, Hoffman, and Blei]{Mandt2017StochasticGD}
S.~Mandt, M.~D. Hoffman, and D.~M. Blei.
\newblock Stochastic gradient descent as approximate {B}ayesian inference.
\newblock \emph{Journal of Machine Learning Research}, 18:\penalty0
  134:1--134:35, 2017.

\bibitem[Martens(2014)]{Martens2014NewIA}
James Martens.
\newblock New insights and perspectives on the natural gradient method.
\newblock 2014.

\bibitem[{Poggio} \& et~al.(2018){Poggio} and et~al.]{2018arXiv180100173P}
T.~{Poggio} and et~al.
\newblock {Theory of Deep Learning III: explaining the non-overfitting puzzle}.
\newblock \emph{ArXiv e-prints, arxive 1801.00173}, 2018.

\bibitem[{Sagun} et~al.(2017){Sagun}, {Evci}, {Ugur Guney}, {Dauphin}, and
  {Bottou}]{2017arXiv170604454S}
L.~{Sagun}, U.~{Evci}, V.~{Ugur Guney}, Y.~{Dauphin}, and L.~{Bottou}.
\newblock {Empirical Analysis of the Hessian of Over-Parametrized Neural
  Networks}.
\newblock \emph{ArXiv e-prints}, 2017.

\bibitem[Saxe et~al.(2018)Saxe, Bansal, Dapello, Advani, Kolchinsky, Tracey,
  and Cox]{michael2018on}
Andrew~Michael Saxe, Yamini Bansal, Joel Dapello, Madhu Advani, Artemy
  Kolchinsky, Brendan~Daniel Tracey, and David~Daniel Cox.
\newblock On the information bottleneck theory of deep learning.
\newblock In \emph{International Conference on Learning Representations}, 2018.
\newblock URL \url{https://openreview.net/forum?id=ry_WPG-A-}.

\bibitem[{Shirish Keskar} et~al.(2016){Shirish Keskar}, {Mudigere}, {Nocedal},
  {Smelyanskiy}, and {Tang}]{2016arXiv160904836S}
N.~{Shirish Keskar}, D.~{Mudigere}, J.~{Nocedal}, M.~{Smelyanskiy}, and
  P.~T.~P. {Tang}.
\newblock {On Large-Batch Training for Deep Learning: Generalization Gap and
  Sharp Minima}.
\newblock \emph{ArXiv e-prints}, 2016.

\bibitem[{Shwartz-Ziv} \& {Tishby}(2017){Shwartz-Ziv} and
  {Tishby}]{2017arXiv170300810S}
R.~{Shwartz-Ziv} and N.~{Tishby}.
\newblock {Opening the Black Box of Deep Neural Networks via Information}.
\newblock \emph{ArXiv e-prints}, March 2017.

\bibitem[Simonyan \& Zisserman(2014)Simonyan and Zisserman]{simonyan2014very}
K.~Simonyan and A.~Zisserman.
\newblock Very deep convolutional networks for large-scale image recognition.
\newblock \emph{arXiv preprint arXiv:1409.1556}, 2014.

\bibitem[{Smith}(2015)]{2015arXiv150601186S}
L.~N. {Smith}.
\newblock {Cyclical Learning Rates for Training Neural Networks}.
\newblock \emph{ArXiv e-prints}, 2015.

\bibitem[Smith \& Le(2017)Smith and Le]{smith2017understanding}
S.L. Smith and Q.V. Le.
\newblock Understanding generalization and stochastic gradient descent.
\newblock \emph{arXiv preprint arXiv:1710.06451}, 2017.

\bibitem[Van~Kampen(1992)]{van1992stochastic}
N.G. Van~Kampen.
\newblock \emph{Stochastic Processes in Physics and Chemistry}.
\newblock North-Holland Personal Library. Elsevier Science, 1992.
\newblock ISBN 9780080571386.
\newblock URL \url{https://books.google.co.uk/books?id=3e7XbMoJzmoC}.

\bibitem[Welling \& Teh(2011)Welling and Teh]{Welling2011}
M.~Welling and Y.~W. Teh.
\newblock Bayesian learning via stochastic gradient {L}angevin dynamics.
\newblock In \emph{Proceedings of the 28th ICML}, pp.\  681--688, 2011.

\bibitem[Wu et~al.(2017)Wu, Zhu, et~al.]{wu2017towards}
Lei Wu, Zhanxing Zhu, et~al.
\newblock Towards understanding generalization of deep learning: Perspective of
  loss landscapes.
\newblock \emph{arXiv preprint arXiv:1706.10239}, 2017.

\bibitem[{Xiao} et~al.(2017){Xiao}, {Rasul}, and
  {Vollgraf}]{2017arXiv170807747X}
H.~{Xiao}, K.~{Rasul}, and R.~{Vollgraf}.
\newblock {Fashion-MNIST: a Novel Image Dataset for Benchmarking Machine
  Learning Algorithms}.
\newblock \emph{ArXiv e-prints}, 2017.

\bibitem[Zhang et~al.(2016)Zhang, Bengio, Hardt, Recht, and
  Vinyals]{zhang2016understanding}
C.~Zhang, S.~Bengio, M.~Hardt, B.~Recht, and Oriol Vinyals.
\newblock Understanding deep learning requires rethinking generalization.
\newblock \emph{arXiv preprint arXiv:1611.03530}, 2016.

\bibitem[Zhang et~al.(2018)Zhang, Saxe, Advani, and
  Lee]{DBLP:journals/corr/abs-1803-01927}
Yao Zhang, Andrew~M. Saxe, Madhu~S. Advani, and Alpha~A. Lee.
\newblock Energy-entropy competition and the effectiveness of stochastic
  gradient descent in machine learning.
\newblock \emph{CoRR}, abs/1803.01927, 2018.
\newblock URL \url{http://arxiv.org/abs/1803.01927}.

\bibitem[{Zhu} et~al.(2018){Zhu}, {Wu}, {Yu}, {Wu}, and
  {Ma}]{2018arXiv180300195Z}
Z.~{Zhu}, J.~{Wu}, B.~{Yu}, L.~{Wu}, and J.~{Ma}.
\newblock {The Regularization Effects of Anisotropic Noise in Stochastic
  Gradient Descent}.
\newblock \emph{ArXiv e-prints}, 2018.

\end{thebibliography}
}

\appendix
\section{When Covariance is Approximately the Hessian}\label{app:CasH}
In this appendix we describe conditions under which the gradient covariance $\mathbf{C}$ can be approximately the same as the Hessian $\mathbf{H}$. 

The covariance matrix $\mathbf{C}$ can be approximated by the sample covariance matrix $\mathbf{K}$, defined in \eqref{KapproxC}.
Define the mean gradient
\begin{align}
\mathbb{E}(\vec{g}_i(\vec{\theta})) = \vec{g}(\vec{\theta}) = \frac{1}{N}\sum_{i=1}^N\vec{g}_i(\vec{\theta})
\end{align}
and the expectation of the squared norm gradient
\begin{align}
\mathbb{E}(\vec{g}_i(\vec{\theta})^T\vec{g}_i(\vec{\theta})) =  {\frac{1}{N} \sum_{i=1}^N \vec{g}_i(\vec{\theta}) ^T\vec{g}_i(\vec{\theta})}
\end{align}

In \citep{michael2018on,2017arXiv170300810S} (see also \citep{2018arXiv180300195Z} who confirm this), they show the squared norm of the mean gradient is much smaller than the expected squared norm of the gradient
\begin{align}
|\vec{g}(\vec{\theta}) |^2 \ll {\frac{1}{N} \sum_{i=1}^N \vec{g}_i(\vec{\theta}) ^T\vec{g}_i(\vec{\theta})}.
\end{align}
From this we have that 
\begin{align}
\vec{g}(\vec{\theta})\vec{g}(\vec{\theta})^T \ll \frac{1}{N} \sum_{i=1}^N \vec{g}_i(\vec{\theta}) \vec{g}_i(\vec{\theta})^T.
\end{align}
We then have that our expression for the sample covariance matrix simplifies to 
\begin{align}
\mathbf{K} \approx \frac{1}{N}  \sum_{i=1}^N \vec{g}_i(\vec{\theta}) \vec{g}_i(\vec{\theta})^T. \label{Casfish}
\end{align}

We follow similar notation to \citep{Martens2014NewIA}. 
Let $f(\vec{x}_i, \vec{\theta})$ be a function mapping the neural network's input to its output. Let $l(\vec{y}, \vec{z})$ be the loss function of an individual sample comparing target $\vec{y}$ to output $\vec{z}$, so we take $\vec{z} = f(\vec{x}_i, \vec{\theta})$ for each sample $i$. Let $P_{\vec{x},\vec{y}}(\vec{\theta})$ be the model distribution, and let $R_{\vec{y}|\vec{z}}$ be the predictive distribution used at the network output, so that $R_{\vec{y}|\vec{z}} = P_{\vec{y}|f(\vec{x}, \vec{\theta})}$.
Let $p_{\vec{\theta}}(\vec{y}|\vec{x})$ be the associated probability density. Many probabilistic models can be formulated by taking the loss function to be 
\begin{align}
l(\vec{y}_i, f(\vec{x}_i, \vec{\theta})) = - \log p_{\vec{\theta}}(\vec{y_i}|\vec{x_i}).
\end{align}
Substituting this into \eqref{Casfish} gives
\begin{align}
\mathbf{K} \approx \frac{1}{N}  \sum_{i=1}^N \frac{\partial \log p_{\vec{\theta}}(\vec{y}_i|\vec{x}_i)}{\partial \vec{\theta}}\frac{\partial \log p_{\vec{\theta}}(\vec{y}_i|\vec{x}_i)}{\partial \vec{\theta}^T} . \label{Caspartial}
\end{align}

Conversely, the Hessian for this probabilistic model can be written as 
\begin{align}
\mathbf{H} = \frac{1}{N}  \sum_{i=1}^N \frac{\partial \log p_{\vec{\theta}}(\vec{y}_i|\vec{x}_i)}{\partial \vec{\theta}}\frac{\partial \log p_{\vec{\theta}}(\vec{y}_i|\vec{x}_i)}{\partial \vec{\theta}^T}  - \frac{1}{p_{\vec{\theta}}(\vec{y}_i|\vec{x}_i)}\frac{\partial^2  p_{\vec{\theta}}(\vec{y}_i|\vec{x}_i)}{\partial \vec{\theta}\partial\vec{\theta}^T}. \label{H}
\end{align}
The first term is the same as appears in the approximation to the sample covariance matrix \eqref{Caspartial}. The second term is negligible
in the case where the model is realizable, i.e. that the {model's conditional} probability distribution coincides with the training {data's conditional} distribution. Mathematically, when the parameter is close to the optimum, $\vec{\theta}_0$, {$p_{\vec{\theta}}(\vec{y}|\vec{x}) = p(\vec{y}|\vec{x}) $}. Under these conditions the model has realized the data distribution and the second term is a sample estimator of the following zero quantity
\begin{align}
\mathbb{E}_{\vec{x}, \vec{y} \sim p(\vec{y},\vec{x})} \left [ \frac{1}{p_{\vec{\theta}}(\vec{y}|\vec{x})}\frac{\partial^2  p_{\vec{\theta}}(\vec{y}|\vec{x})}{\partial \vec{\theta}\partial\vec{\theta}^T}\right ] 
&= \int d\vec{x}d\vec{y}  p(\vec{x})\frac{\partial^2  p_{\vec{\theta}}(\vec{y}|\vec{x})}{\partial \vec{\theta}\partial\vec{\theta}^T}  
\\
& = \int d\vec{x} p(\vec{x})\frac{\partial^2}{\partial \vec{\theta}\partial\vec{\theta}^T}\left[\int  d\vec{y}  \  p_{\vec{\theta}}(\vec{y}|\vec{x})\right] 
\\
& = \int d\vec{x} p(\vec{x}) \frac{\partial^2}{\partial \vec{\theta}\partial\vec{\theta}^T}\left[1\right]   = 0 ,
\end{align}
with the estimator becoming more accurate with larger $N$.
Thus we have that the covariance is approximately the Hessian\footnote{We also note that the first term is the same as the Empirical Fisher. The same argument can be used \citep{Martens2014NewIA} to demonstrate that the Empirical Fisher matrix approximates the Hessian, and that Natural Gradient \citep{Amari:1998:NGW:287476.287477} close to the optimum is similar to the Newton method.}.

\end{document}